\documentclass{article}

\usepackage{arxiv}

\usepackage[utf8]{inputenc}
\usepackage[T1]{fontenc}
\usepackage{hyperref}
\usepackage{url}
\usepackage{booktabs}
\usepackage{amsfonts}
\usepackage{nicefrac}
\usepackage{microtype}
\usepackage{lipsum}
\usepackage{graphicx}
\usepackage{natbib}
\usepackage{doi}
\usepackage{amsmath,bm}

\def\eqref#1{equation~\ref{#1}}

\def\1{\bm{1}}

\def\rx{{\textnormal{x}}}

\def\rz{{\textnormal{z}}}

\def\rtheta{{\theta}}

\def\rvw{{\mathbf{w}}}
\def\rvx{{\mathbf{x}}}

\def\rvz{{\mathbf{z}}}

\def\rvtheta{{\bm{\theta}}}

\def\ervx{{\textnormal{x}}}

\def\ervz{{\textnormal{z}}}

\def\ervtheta{{\theta}}

\def\rmW{{\mathbf{W}}}
\def\rmX{{\mathbf{X}}}

\def\rmZ{{\mathbf{Z}}}

\def\rmTheta{{\bm{\Theta}}}

\def\vc{{\bm{c}}}

\def\vx{{\bm{x}}}

\def\valpha{{\bm{\alpha}}}
\def\vgamma{{\bm{\gamma}}}

\def\vpi{{\bm{\pi}}}

\def\evalpha{{\alpha}}

\def\evgamma{{\gamma}}

\def\evc{{c}}

\DeclareMathAlphabet{\mathsfit}{\encodingdefault}{\sfdefault}{m}{sl}
\SetMathAlphabet{\mathsfit}{bold}{\encodingdefault}{\sfdefault}{bx}{n}

\newcommand{\childs}{Ch}

\DeclareMathOperator*{\argmax}{arg\,max}

\usepackage{makecell}

\usepackage{cleveref}
\usepackage{xcolor}

\definecolor{colorblue}{HTML}{469FF8}
\definecolor{colorred}{HTML}{EC615C}
\definecolor{colorgreen}{HTML}{81D652}

\newcommand{\good}[1]{\textcolor{colorgreen}{#1}}
\newcommand{\bad}[1]{\textcolor{colorred}{#1}}
\usepackage{algorithm}
\usepackage{algpseudocode}
\usepackage{arxiv}
\usepackage{cleveref}
\usepackage[inline]{enumitem}
\usepackage{physics}
\usepackage{stfloats}
\usepackage{subcaption}
\usepackage{tabularx}
\usepackage{tikz}
\usetikzlibrary{bayesnet}
\usetikzlibrary{arrows}
\usetikzlibrary{calc, positioning}
\newcommand{\summark}{$+$}
\newcommand{\prodmark}{$\times$}
\newcommand{\leafmark}{\includegraphics[width=0.5cm, clip]{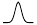}}

\title{Top-Down Bayesian Posterior Sampling for Sum-Product Networks}

\date{June 17, 2024}

\newif\ifuniqueAffiliation
\uniqueAffiliationtrue

\ifuniqueAffiliation
\author{Soma Yokoi \\
    Department of Computer Science\\
	The University of Tokyo\\
	\texttt{syokoi@g.ecc.u-tokyo.ac.jp} \\
	\And
	Issei Sato \\
	Department of Computer Science\\
	The University of Tokyo\\
	\texttt{sato@g.ecc.u-tokyo.ac.jp} \\
}
\else
\usepackage{authblk}

\author[1]{David S.~Hippocampus\thanks{\texttt{hippo@cs.cranberry-lemon.edu}}}
\author[1,2]{Elias D.~Striatum\thanks{\texttt{stariate@ee.mount-sheikh.edu}}}
\affil[1]{Department of Computer Science, Cranberry-Lemon University, Pittsburgh, PA 15213}
\affil[2]{Department of Electrical Engineering, Mount-Sheikh University, Santa Narimana, Levand}
\fi

\renewcommand{\shorttitle}{Top-Down Bayesian Posterior Sampling for Sum-Product Networks}

\hypersetup{
pdftitle={Top-Down Bayesian Posterior Sampling for Sum-Product Networks},
pdfsubject={},
pdfauthor={Soma Yokoi, Issei Sato},
pdfkeywords={Sum-Product Network, Bayesian inference, Gibbs Sampling},
}

\begin{document}
\maketitle

\begin{abstract}
Sum-product networks (SPNs) are probabilistic models characterized by exact and fast evaluation of fundamental probabilistic operations.
Its superior computational tractability has led to applications in many fields, such as machine learning with time constraints or accuracy requirements and real-time systems.
The structural constraints of SPNs supporting fast inference, however, lead to increased learning-time complexity and can be an obstacle to building highly expressive SPNs.
This study aimed to develop a Bayesian learning approach that can be efficiently implemented on large-scale SPNs.
We derived a new full conditional probability of Gibbs sampling by marginalizing multiple random variables to expeditiously obtain the posterior distribution.
The complexity analysis revealed that our sampling algorithm works efficiently even for the largest possible SPN.
Furthermore, we proposed a hyperparameter tuning method that balances the diversity of the prior distribution and optimization efficiency in large-scale SPNs.
Our method has improved learning-time complexity and demonstrated computational speed tens to more than one hundred times faster and superior predictive performance in numerical experiments on more than $20$ datasets.
\end{abstract}

\keywords{Sum-Product Network, Bayesian inference, Gibbs Sampling}

\maketitle

\section{Introduction}
Probabilistic machine learning can account for uncertainty, but many important inference tasks often encounter computational difficulties due to high-dimensional integrals.
Simple probabilistic models, such as factorized and mixture models, can compute fundamental probabilistic operations exactly in polynomial time to the model size, e.g., likelihood, marginalization, conditional distribution, and moments as shown in \Cref{tab:SPN_model_tractability}. 
However, these models often suffer from insufficient expressive power due to the lack of scalability to complex structures.
Consequently, the primary focus of current probabilistic modeling is to achieve both expressiveness and computational tractability.

A sum-product network (SPN)~\citep{Poon11} has received much attention in recent years thanks to its tractability by design.
SPNs are considered a deep extension of the factorized and mixture models while maintaining their tractability.
The tractability of SPNs is a notable characteristic, markedly distinct from models using deep neural networks (DNNs) such as generative adversarial networks~\citep{Goodfellow14}, variational autoencoders~\citep{Kingma14}, and normalizing flows~\citep{Rezende15}, where many operations are approximated.
SPNs have many possible uses for machine learning tasks that require fast and exact inference, e.g., image segmentation~\citep{Yuan16, Rathke17}, speech processing~\citep{Peharz14}, language modeling~\citep{Cheng14}, and cosmological simulations~\citep{Parag22}.
SPNs have also been investigated for real-time systems such as activity recognition~\citep{Amer16}, robotics~\citep{Pronobis17}, probabilistic programming~\citep{Saad21}, and hardware design~\citep{Sommer18}.

Learning an SPN is relatively more complicated and time consuming than inference. 
Conventional approaches like gradient descent~\citep{Poon11,Gens12} and an expectation-maximization algorithm~\citep{Poon11,Hsu17} often suffer from overfitting, mode collapse, and instability in missing data.
One way to prevent these problems is Bayesian learning.
Bayesian moment matching~\citep{Jaini16} and collapsed variation inference~\citep{zhaoCollapsedVariationalInference2016} were proposed, and in recent years, Gibbs sampling has been remarkable for generating samples from the posterior distribution.
\citet{Vergari19} obtained the samples from the full conditional distribution using ancestral sampling that recursively traces the graph from the leaf nodes toward the root, and \citet{Trapp19} showed that the model parameters and the network structure can be learned simultaneously in a theoretically consistent Bayesian framework.
However, although these studies validated the flexibility of Bayesian learning, little attention has been paid to the computational complexity.

The structural constraints of sum and product nodes guarantee the inference-time tractability of SPNs but lead to increased time complexity during posterior sampling.
The scope of a node, typically feature dimensions included in its children, is constrained by \textit{decomposability} and \textit{completeness} conditions.
SPNs still possess high expressiveness despite these constraints, although the graph shape is restricted.
In particular, the height of SPNs has a strict upper limit, so the graph tends to be broad.
\citet{Ko2020} reported that existing structural learning methods~\citep{pmlr-v28-gens13,10.5555/3120406.3120431,zhaoUnifiedApproachLearning2016,Butz17} are more likely to generate wide graphs, which can be a performance bottleneck for inference.
It is also computationally disadvantageous for existing posterior sampling that relies on bottom-up traversal of the entire graph.
This problem is currently an obstacle to building highly expressive large-scale SPNs.

This study aims to extend the efficiency of SPNs to Bayesian learning.
To enable Bayesian learning of SPNs in a reasonable time for large network sizes, we solve the three most critical problems from theory to practice:
\begin{enumerate*}
  \item A new full conditional probability of Gibbs sampling is derived by marginalizing multiple random variables to obtain the posterior distribution.
  \item To efficiently sample from the derived probability, we propose a novel sampling algorithm, \textit{top-down sampling} algorithm, based on the Metropolis--Hastings method considering the complexity of SPNs.
  It reduces the exponent by one in learning-time complexity and achieves tens to more than one hundred times faster runtime in numerical experiments.
  \item For the increasing hyperparameters of large-scale SPNs, a new tuning method is proposed following an empirical Bayesian approach.
\end{enumerate*}

This paper is structured as follows. 
\Cref{sec:SPN_background} quickly reviews SPNs.
We describe an existing latent variable model and Gibbs sampling and illustrate how they are impaired by large-scale SPNs.
\Cref{sec:SPN_propose} reveals the complexity of SPNs and introduces our marginalized posterior, fast sampling algorithm, and hyperparameter tuning method and discusses their theoretical computational complexity. 
\Cref{sec:SPN_experiment} demonstrates the computational speed, sample correlation, and predictive performance of the proposed method through numerical experiments on more than $20$ datasets.
Finally, \Cref{sec:SPN_conclusion} concludes the study.

\begin{table}[t]
    \caption{Tractability of Probabilistic Models \citep{robertpeharzSumProductNetworksDeep,Trapp20}}
    \label{tab:SPN_model_tractability}
    \centering
      \begin{tabular}{ccccc}
          \toprule
                      & Factorized   & Mixture      & \textbf{SPN}          & DNN \\
          \midrule
          Sampling    & $\good{\bm\checkmark}$ & $\good{\bm\checkmark}$ & $\good{\bm\checkmark}$ & $\good{\bm\checkmark}$ \\
          Density     & $\good{\bm\checkmark}$ & $\good{\bm\checkmark}$ & $\good{\bm\checkmark}$ & $\good{\bm\checkmark}/\bad{\bm\times}$ \\
          Marginal    & $\good{\bm\checkmark}$ & $\good{\bm\checkmark}$ & $\good{\bm\checkmark}$ & $\bad{\bm\times}$ \\
          Conditional & $\good{\bm\checkmark}$ & $\good{\bm\checkmark}$ & $\good{\bm\checkmark}$ & $\bad{\bm\times}$ \\
          Moment      & $\good{\bm\checkmark}$ & $\good{\bm\checkmark}$ & $\good{\bm\checkmark}$ & $\bad{\bm\times}$ \\
          \midrule
          Deep structure & $\bad{\bm\times}$ & $\bad{\bm\times}$ & $\good{\bm\checkmark}$ & $\good{\bm\checkmark}$ \\
          \bottomrule
      \end{tabular}
  \end{table}
  
  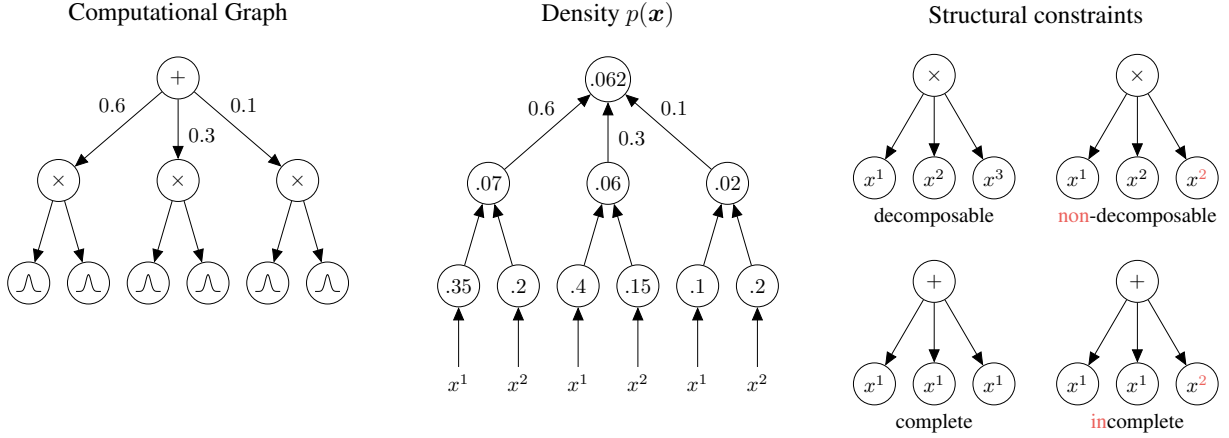
\begin{figure*}[t]
    \centering
    \resizebox{\linewidth}{!}{
        \begin{tikzpicture}
            \node (a) {\large Computational Graph};
            \node [right=4cm of a] (b) {\large Density $p(\vx)$};
            \node [right=4cm of b] (c) {\large Structural constraints};
          
            \begin{scope}[shift={($(a.south) + (0cm, -0.75cm)$)}]
              \node[latent] (s1) {\summark};
              \node[latent,below=of s1,xshift=-2cm] (p1) {\prodmark};
              \node[latent,below=of s1] (p2) {\prodmark};
              \node[latent,below=of s1,xshift=2cm] (p3) {\prodmark};
              \node[latent,below=of p1,xshift=-0.5cm] (d1) {\leafmark};
              \node[latent,below=of p1,xshift=+0.5cm] (d2) {\leafmark};
              \node[latent,below=of p2,xshift=-0.5cm] (d3) {\leafmark};
              \node[latent,below=of p2,xshift=+0.5cm] (d4) {\leafmark};
              \node[latent,below=of p3,xshift=-0.5cm] (d5) {\leafmark};
              \node[latent,below=of p3,xshift=+0.5cm] (d6) {\leafmark};
              \path[->] (s1) edge node[shift={(-0.1cm, 0.4cm)}] {$0.6$} (p1);
              \path[->] (s1) edge node[shift={(+0.4cm, -0.1cm)}] {$0.3$} (p2);
              \path[->] (s1) edge node[shift={(+0.1cm, 0.4cm)}] {$0.1$} (p3);
              \edge {p1} {d1,d2}
              \edge {p2} {d3,d4}
              \edge {p3} {d5,d6}
            \end{scope}
          
            \begin{scope}[shift={($(b.south) + (0cm, -0.75cm)$)}]
              \node[latent] (s1) {$.062$};
              \node[latent,below=of s1,xshift=-2cm] (p1) {$.07$};
              \node[latent,below=of s1] (p2) {$.06$};
              \node[latent,below=of s1,xshift=2cm] (p3) {$.02$};
              \node[latent,below=of p1,xshift=-0.5cm] (d1) {$.35$};
              \node[latent,below=of p1,xshift=+0.5cm] (d2) {$.2$};
              \node[latent,below=of p2,xshift=-0.5cm] (d3) {$.4$};
              \node[latent,below=of p2,xshift=+0.5cm] (d4) {$.15$};
              \node[latent,below=of p3,xshift=-0.5cm] (d5) {$.1$};
              \node[latent,below=of p3,xshift=+0.5cm] (d6) {$.2$};
              \node[below=of d1] (x1) {$x^1$};
              \node[below=of d2] (x2) {$x^2$};
              \node[below=of d3] (x3) {$x^1$};
              \node[below=of d4] (x4) {$x^2$};
              \node[below=of d5] (x5) {$x^1$};
              \node[below=of d6] (x6) {$x^2$};
              \path[->] (p1) edge node[shift={(-0.1cm, 0.4cm)}] {$0.6$} (s1);
              \path[->] (p2) edge node[shift={(+0.4cm, -0.1cm)}] {$0.3$} (s1);
              \path[->] (p3) edge node[shift={(+0.1cm, 0.4cm)}] {$0.1$} (s1);
              \edge {d1,d2} {p1}
              \edge {d3,d4} {p2}
              \edge {d5,d6} {p3}
              \edge {x1} {d1}
              \edge {x2} {d2}
              \edge {x3} {d3}
              \edge {x4} {d4}
              \edge {x5} {d5}
              \edge {x6} {d6}
            \end{scope}
          
            \begin{scope}[shift={($(c.south) + (-1.7cm, -0.75cm)$)}]
              \node[latent] (p1) {\prodmark};
              \node[latent,below=of p1,xshift=-1cm] (d1) {$x^1$};
              \node[latent,below=of p1] (d2) {$x^2$};
              \node[latent,below=of p1,xshift=1cm] (d3) {$x^3$};
              \edge {p1} {d1,d2,d3}
              \node[shift={($(d2.south) + (0cm, -0.3cm)$)}] (n1) {decomposable};
            \end{scope}
          
            \begin{scope}[shift={($(c.south) + (+1.7cm, -0.75cm)$)}]
              \node[latent] (p2) {\prodmark};
              \node[latent,below=of p2,xshift=-1cm] (d4) {$x^1$};
              \node[latent,below=of p2] (d5) {$x^2$};
              \node[latent,below=of p2,xshift=1cm] (d6) {$x^{\bad{2}}$
              };
              \edge {p2} {d4,d5,d6}
              \node[shift={($(d5.south) + (0cm, -0.3cm)$)}] (n2) {\bad{non}-decomposable};
            \end{scope}
          
            \begin{scope}[shift={($(c.south) + (-1.7cm, -4.2cm)$)}]
              \node[latent] (s1) {\summark};
              \node[latent,below=of s1,xshift=-1cm] (d7) {$x^1$};
              \node[latent,below=of s1] (d8) {$x^1$};
              \node[latent,below=of s1,xshift=1cm] (d9) {$x^1$};
              \edge {s1} {d7,d8,d9}
              \node[shift={($(d8.south) + (0cm, -0.3cm)$)}] (n3) {complete};
            \end{scope}
          
            \begin{scope}[shift={($(c.south) + (+1.7cm, -4.2cm)$)}]
              \node[latent] (s2) {\summark};
              \node[latent,below=of s2,xshift=-1cm] (d10) {$x^1$};
              \node[latent,below=of s2] (d11) {$x^1$};
              \node[latent,below=of s2,xshift=1cm] (d12) {$x^{\bad{2}}$};
              \edge {s2} {d10,d11,d12}
              \node[shift={($(d11.south) + (0cm, -0.3cm)$)}] (n4) {\bad{in}complete};
            \end{scope}
          
            \draw(0,0)node[shift={($(b.south) - (0, 1.0cm)$)}] (center) {};
            \path (center) -- ++($-1*($(current bounding box.west |- center) - (center)$)$)
                  (center) -- ++($-1*($(current bounding box.east |- center) - (center)$)$);
        \end{tikzpicture}
    }
    \caption{Overview of SPNs. Left: computational graph representing the structure and weights of the SPN. Center: evaluation of the density of input data $\vx$. Right: structural constraints ensuring tractability.}
    \label{fig:SPN_overview}
\end{figure*}

\section{Bayesian Sum-Product Networks}\label{sec:SPN_background}
This section quickly reviews SPNs and introduces their basics and constraints.
We also describe an existing latent variable model and approximate Bayesian learning by Gibbs sampling and illustrate how they are impaired by large-scale SPNs.

\subsection{Sum-Product Network}
An SPN is a probabilistic model with a computational graph represented as a rooted directed acyclic graph (DAG) composed of sum, product, distribution nodes, and edges.
Figure~\ref{fig:SPN_overview} left illustrates a typical computational graph of an SPN. 
The product node corresponds to the factorized model, i.e., a product of probability distributions $p(\vx) = \prod_d p_d(x^d)$, and the sum node is the mixture model, i.e., a weighted sum of probability distributions $p(\vx) = \sum_k w_k p_k(\vx)$.
These nodes exist at the root and intermediate levels, performing the operations on their children's outputs and producing results. 
SPNs can stack factorized and mixture models in a complex manner by repeating the product and sum nodes alternately.
The distribution node at the leaves consists of a probability distribution and its parameters for a simple, typically one-dimensional variable.

The modeling probability of SPNs corresponds to the bottom-up evaluation on the computational graph.
Figure~\ref{fig:SPN_overview} center depicts an example of evaluating density $p(\vx)$ of $D=2$ input data $\vx=(x^1,x^2)$:
\begin{enumerate*}
    \item The input vector $\vx$ is divided into simple variables $x^1$ and $x^2$ and input to the corresponding distribution nodes.
    The input value is evaluated by the probability of the distribution node and outputs a scalar probability value.
    \item The intermediate node that receives the children's output computes their convex combination or product and outputs the scalar value.
    This procedure is repeated bottom-up toward the root.
    \item All leaves and intermediate nodes are computed, and the final output value of the SPN is computed at the root node.
\end{enumerate*}

Two simple constraints on the structure of the graph streamline the fundamental probabilistic operations during inference.
Figure~\ref{fig:SPN_overview} right represents the structural constraints of SPNs.
The children of product nodes must have mutually exclusive variables, called \emph{decomposability}.  
Decomposability simplifies the integration of a product node into the integration of its children.
The children of sum nodes must include common variables, called \emph{completeness}.
Completeness ensures efficient model counting and minimization of cardinality~\citep{Darwiche2001-DAROTT-2} and simplifies the integration of a sum node into the integration of its children.
SPNs satisfying both conditions can compute any marginalization in linear time with respect to the number of nodes~\citep{cs6790}.
The exact evaluation of density, marginalization, conditional probability, and moment~\citep{10.5555/3295222.3295433} can be performed efficiently.

\subsection{Latent Variable Model}\label{sec:background_latent}
An SPN can be interpreted within the Bayesian framework by considering it as a latent variable model.
Let us introduce a categorical latent variable $\rz$ that indicates a mixture component of the sum node.
The latent state $\rvz=(\rz^1,\ldots,\rz^S)$ of the SPN is determined by specifying the state $\rz^s=c$ where $c\in(1,\ldots,C_s)$ for each sum node $s\in(1,\ldots,S)$.

A multinomial distribution is considered as the underlying probability distribution for latent variable $\ervz^s$, and a Dirichlet distribution is given as the conjugate prior distribution
\begin{equation}\label{equ:SPN_modeling_z}
  \begin{aligned} 
    \ervz^s_n &\sim \operatorname{Multi}\qty(\ervz^s_n \mid \rvw^s) \ \forall s \ \forall n, &
    \rvw^s &\sim \operatorname{Dir}\qty(\rvw^s \mid \evalpha^s) \ \forall s.
  \end{aligned}
\end{equation}
The probability distribution of the distribution node is given by a distribution $L_j^d$ parameterized with $\theta_j^d$ and its corresponding conjugate prior distribution
\begin{equation}\label{equ:SPN_modeling_x}
  \begin{aligned}
    \vx_n &\sim \prod_{L_j^d \in T\qty(\rvz_n)} L_j^d\qty(\vx_n^d \mid \rvtheta_j^d) \ \forall n, &
    \ervtheta_j^d &\sim p\qty(\ervtheta_j^d \mid \vgamma_j^d) \ \forall j \ \forall d.
  \end{aligned}
\end{equation}
The assumption of conjugacy is inherited from existing studies\,\citep{Vergari19,Trapp19}, and it is noted that even in non-conjugate cases, one-dimensional leaf distributions can be easily approximated numerically.
The probability distribution $L_j^d$ is typically chosen by the variable type (continuous or discrete) and its domain (real numbers, positive numbers, finite interval $[0,1]$, etc.).
Alternatively, the model can automatically choose the appropriate distribution by configuring the distribution node as a heterogeneous mixture distribution~\citep{Vergari19}.
The dependency between the random variables and hyperparameters is summarized in \Cref{fig:SPN_Bayes} left.

As indicated by the bold edges in \Cref{fig:SPN_Bayes} center and right, a subgraph from the root to the leaves is obtained by removing the unselected edges from the graph.
This is called an \emph{induced tree}~\citep{zhaoUnifiedApproachLearning2016}.
We denote the induced tree determined by the state $\rvz$ as $T(\rvz)$.
The induced tree always includes one distribution node for each feature dimension, $L_j^d \in T(\rvz)  \ \ \forall d \in (1,\ldots,D)$, due to the decomposability and completeness.

\subsection{Gibbs Sampling}
From the plate notation in \Cref{fig:SPN_Bayes} left, the posterior distribution of SPNs can be obtained as
\begin{equation}\begin{split}\label{equ:previous_posterior}
    p\qty(\rmZ, \rmW, \rmTheta \mid \rmX, \valpha, \vgamma) 
    \propto & \ p\qty(\rmX \mid \rmZ, \rmTheta) p\qty(\rmZ \mid \rmW) p\qty(\rmW \mid \valpha) p\qty(\rmTheta \mid \vgamma).
\end{split}\end{equation}

In Bayesian learning of SPNs, it is necessary to numerically realize the posterior distribution.
Previous studies~\citep{Vergari19,Trapp19} have used Gibbs sampling to generate samples alternately from the conditional probabilities of the three random variables $\rmZ$, $\rmW$, and $\rmTheta$:
\begin{equation}\label{equ:SPN_previous_conditional}
    p\qty(\rvz_n \mid \rvx_n, \rmTheta, \rmW) \ \forall n, \ \ \ p\qty(\rmTheta \mid \rmX, \rmZ), \ \ \ p\qty(\rmW \mid \rmZ).
\end{equation}
Since it is difficult to sample all categorical variables $\rmZ$ at once, they are divided into $\rvz_n \ \forall n\in(1,\ldots,N)$.

Ancestral sampling is employed to generate the $S$-dimensional latent vector $\rvz_n = (\rz_n^1,\ldots,\rz_n^S)$ from the joint distribution.
Starting at the root node, sample the branch $c\in\childs(s)$ for each sum node $s$ encountered
\begin{equation}
    p_s\qty(\rz_n^s = c \mid \rvx_n, \rmTheta, \rmW) \propto w_{s,c} \ p_c\qty(\rvx_n\mid\rmTheta, \rmW).
\end{equation}
\Cref{fig:SPN_Bayes} center illustrates joint sampling of latent vector $\rvz_n$ using ancestral sampling.
The algorithm runs bottom-up, where the output is determined first \citep{Poon11,Vergari19,Trapp19}.
The states excluded from the induced tree are sampled from the prior \citep{10.1109/TPAMI.2016.2618381}.

Posterior sampling using this approach is often computationally challenging for large SPNs.
Ancestral sampling requires a traversal of the entire graph, which is too costly to be executed inside the most critical loop of Gibbs sampling.
The larger SPNs typically require more iterations to reach a stationary state, making the problem more serious.

\begin{figure*}[t]
    \centering
    \resizebox{\linewidth}{!}{
        \begin{tikzpicture}
            \node (a) {\large Plate Notation};
            \node [right=5.5cm of a] (b) {\large Bottom-up Sampling};
            \node [right=5.5cm of b] (c) {\large Top-down Sampling};
          
            \begin{scope}[shift={($(a.south) - (0, 1.5cm)$)}]
                \node[latent,yshift=-5.0cm] (theta) {$\rvtheta_j^d$};
                \node[obs,left=of theta] (x) {$\rx_n^d$};
                \node[circle,right=of theta] (gamma) {$\vgamma_j^d$};
                \node[latent,above=of x] (z) {$\rz_n^s$};
                \node[latent,above=of z] (w) {$\rvw^s$};
                \node[circle,above=of w] (alpha) {$\valpha^s$};
                \draw (-2.75,-5.75) rectangle (-0.75,-2.25);
                \node at (-2.5,-5.5) {$N$};
                \draw (-0.5,-5.75) rectangle (2.5,-4.25);
                \node at (2.25,-5.5) {$J^d$};
                \draw (-3,-6.0) rectangle (3,-4);
                \node at (2.75,-5.75) {$D$};
                \draw (-3,-3.75) rectangle (-0.5,1.0);
                \node at (-0.75, 0.75) {$S$};
                \edge {alpha} {w}
                \edge {w} {z}
                \edge {z} {x}
                \edge {theta} {x}
                \edge {gamma} {theta}
            \end{scope}
          
            \begin{scope}[shift={($(b.south) - (0, 1.0cm)$)}]
              \draw(2.0cm,0.5cm)[colorred] node{$\rvz=(2,*,*,1,2)$};
              \node[latent] (s1) {\summark};
              \node[latent,below=of s1,xshift=-2cm] (p1) {\prodmark};
              \node[latent,below=of s1,xshift=2cm] (p2) {\prodmark};
              \node[latent,below=of p1,xshift=-1cm] (s2) {\summark};
              \node[latent,below=of p1,xshift=1cm] (s3) {\summark};
              \node[latent,below=of p2,xshift=-1cm] (s4) {\summark};
              \node[latent,below=of p2,xshift=1cm] (s5) {\summark};
              \node[latent,below=of s2,xshift=-0.5cm] (d1) {\leafmark};
              \node[latent,below=of s2,xshift=0.5cm] (d2) {\leafmark};
              \node[latent,below=of s3,xshift=-0.5cm] (d3) {\leafmark};
              \node[latent,below=of s3,xshift=0.5cm] (d4) {\leafmark};
              \node[latent,below=of s4,xshift=-0.5cm] (d5) {\leafmark};
              \node[latent,below=of s4,xshift=0.5cm] (d6) {\leafmark};
              \node[latent,below=of s5,xshift=-0.5cm] (d7) {\leafmark};
              \node[latent,below=of s5,xshift=0.5cm] (d8) {\leafmark};
              \node[below=of d1] (x1) {$x^1$};
              \node[below=of d2] (x2) {$x^1$};
              \node[below=of d3] (x3) {$x^2$};
              \node[below=of d4] (x4) {$x^2$};
              \node[below=of d5] (x5) {$x^1$};
              \node[below=of d6] (x6) {$x^1$};
              \node[below=of d7] (x7) {$x^2$};
              \node[below=of d8] (x8) {$x^2$};
              \edge[dotted] {p1} {s1}
              \path[->,ultra thick, colorred] (p2) edge node[shift={(0.8cm, 0.1cm)}] {$\rz^1=2$} (s1);
              \edge {s2, s3} {p1}
              \edge[ultra thick] {s4, s5} {p2}
              \path[->, colorred] (d1) edge node[shift={(0.6cm, 0.0cm)}] {$\rz^2=1$} (s2);
              \edge[dotted] {d2} {s2}
              \path[->, colorred] (d3) edge node[shift={(0.6cm, 0.0cm)}] {$\rz^3=1$} (s3);
              \edge[dotted] {d4} {s3}
              \path[->,ultra thick, colorred] (d5) edge node[shift={(0.6cm, 0.0cm)}] {$\rz^4=1$} (s4);
              \edge[dotted] {d6} {s4}
              \edge[dotted] {d7} {s5}
              \path[->,ultra thick, colorred] (d8) edge node[shift={(0.6cm, 0.0cm)}] {$\rz^5=2$} (s5);
              \edge {x1} {d1}
              \edge {x2} {d2}
              \edge {x3} {d3}
              \edge {x4} {d4}
              \edge {x5} {d5}
              \edge {x6} {d6}
              \edge {x7} {d7}
              \edge {x8} {d8}
            \end{scope}

            \begin{scope}[shift={($(c.south) - (0, 1.0cm)$)}]
                \draw(2.0cm,0.5cm)[colorgreen] node{$\widehat{\rvz}=(2,1,1,1,2)$};
                \node[latent] (s1) {\summark};
                \node[latent,below=of s1,xshift=-2cm,dotted] (p1) {\prodmark};
                \node[latent,below=of s1,xshift=2cm] (p2) {\prodmark};
                \node[latent,below=of p1,xshift=-1cm,dotted] (s2) {\summark};
                \node[latent,below=of p1,xshift=1cm,dotted] (s3) {\summark};
                \node[latent,below=of p2,xshift=-1cm] (s4) {\summark};
                \node[latent,below=of p2,xshift=1cm] (s5) {\summark};
                \node[latent,below=of s2,xshift=-0.5cm,dotted] (d1) {\leafmark};
                \node[latent,below=of s2,xshift=0.5cm,dotted] (d2) {\leafmark};
                \node[latent,below=of s3,xshift=-0.5cm,dotted] (d3) {\leafmark};
                \node[latent,below=of s3,xshift=0.5cm,dotted] (d4) {\leafmark};
                \node[latent,below=of s4,xshift=-0.5cm] (d5) {\leafmark};
                \node[latent,below=of s4,xshift=0.5cm,dotted] (d6) {\leafmark};
                \node[latent,below=of s5,xshift=-0.5cm,dotted] (d7) {\leafmark};
                \node[latent,below=of s5,xshift=0.5cm] (d8) {\leafmark};
                \node[below=of d5] (x1) {$x^1$};
                \node[below=of d8] (x2) {$x^2$};
                \path[dotted] (s1) edge (p1);
                \path[->,ultra thick, colorgreen] (s1) edge node[shift={(0.8cm, 0.1cm)}] {$\widehat{\rz}^1=2$} (p2);
                \path[dotted] (p1) edge (s2);
                \path[dotted] (p1) edge (s3);
                \edge[ultra thick] {p2} {s4, s5}
                \path[dotted] (s2) edge (d1);
                \path[dotted] (s2) edge (d2);
                \path[dotted] (s3) edge (d3);
                \path[dotted] (s3) edge (d4);
                \path[->,ultra thick, colorgreen] (s4) edge node[shift={(0.6cm, 0.1cm)}] {$\widehat{\rz}^4=1$} (d5);
                \path[dotted] (s4) edge (d6);
                \path[dotted] (s5) edge (d7);
                \path[->,ultra thick, colorgreen] (s5) edge node[shift={(0.6cm, 0.1cm)}] {$\widehat{\rz}^5=2$} (d8);
                \edge {d5} {x1};
                \edge {d8} {x2};
            \end{scope}

            \draw(0,0)node[shift={($(b.south) - (0, 1.0cm)$)}] (center) {};
            \path (center) -- ++($-1*($(current bounding box.west |- center) - (center)$)$)
                  (center) -- ++($-1*($(current bounding box.east |- center) - (center)$)$);
        \end{tikzpicture}
    }
    \caption{Bayesian SPNs. Left: plate notation showing the conditional dependencies of random variables based on the Bayesian interpretation of SPNs. Center: entire graph traversal for ancestral sampling in the bottom-up approach. Right: subgraph access for rejection sampling in the proposed top-down approach.}
    \label{fig:SPN_Bayes}
\end{figure*}

\section{Proposed Method}\label{sec:SPN_propose}
This section reveals the complexity of SPNs and the structures that highly expressive SPNs tend to form.
Then we propose a novel posterior sampling method for the SPNs and discuss its theoretical advantages. 
The proposed method is explained from three perspectives: a new full conditional probability of Gibbs sampling that marginalizes multiple random variables, a top-down sampling algorithm that includes new proposal and rejection steps, and hyperparameter tuning using an empirical Bayesian approach.

\subsection{Complexity of SPNs}\label{sec:SPN_complexity}

Both top-down and bottom-up algorithms are applicable to DAG-structured SPNs.
However, the ensuing discussion on computational complexity confines SPNs to tree structures.
Tree-structured SPNs are not necessarily compact representations~\citep{Trapp20}, but they are frequently used in previous studies~\citep{pmlr-v28-gens13,10.5555/3120406.3120431,Vergari19,Trapp19,Trapp2019Optim,pmlr-v115-peharz20a}.

This study aims to propose a sampling method that efficiently operates even on SPNs of theoretically maximum size. 
For computational considerations, we assume that SPNs fulfill the following conditions: 
\begin{enumerate*}
    \item They satisfy the structural constraints of completeness and decomposability. 
    \item The outdegrees of sum and product nodes are $C_s, C_p \geq 2$, respectively. 
    \item The graph exhibits a tree structure, implying that the nodes do not have overlapping children. 
    \item The root is a sum node.
\end{enumerate*}
These assumptions enable subsequent algorithmic discussions to cover the worst-case computational complexity under given $D$, $C_s$, and $C_p$.

The structural constraints are important for making SPNs tractable for fundamental probability operations, but they restrict the graph shape.
\Cref{tab:SPN_model_complexity} shows the maximum possible size of an SPN. 
For simplicity, it only presents cases where the SPN is a complete tree, i.e., the product nodes evenly distribute children, resulting in $\log_{C_p}D$ being an integer, or conversely, where the children of product nodes are maximally skewed, with $\frac{D-1}{C_p-1}$ being an integer. 
Other tree structures fall within the intermediate of these cases.
In either case, the number of all nodes $V$, distribution nodes $L$, induced trees, and graph breadth are larger than the other variables.
Note that the sizes may differ if one considers SPNs with different conditions.
For example, $V$ and $S$ are proportional in \citet{zhaoCollapsedVariationalInference2016}, while they are asymptotically different by a degree of one in our case.

The height and breadth of the SPN should be noted.
Increasing $C_s$ can widen the graph, whereas increasing $C_p$ reduces the height.
Since the product nodes must have children of different feature dimensions for decomposability, the number of product nodes is restricted by $D$.
Each time a product node is passed through on the graph, it \textit{consumes} $C_p$ dimensions from $D$.
When the SPN is a complete tree, the product node satisfying decomposability can only be used up to $\log_{C_p}D$ times from a leaf to the root.
For the SPN where sum and product nodes alternate, the graph height is limited to $2\log_{C_p}D + 2$, including the root and leaves.
When the consumption is minimized, i.e., $C_p=2$, the graph height is maximized to $2\log_2D + 2$.
In contrast, the outdegree $C_s$ of sum nodes can be increased while satisfying completeness.
Accordingly, the graph breadth can increase as $C_s\cdot\qty(C_p C_s)^{\log_{C_p}D}$ without limitation.

\begin{figure}
    \begin{subfigure}{0.49\linewidth}
        \centering
        \includegraphics[keepaspectratio, width=\linewidth, height=10cm]{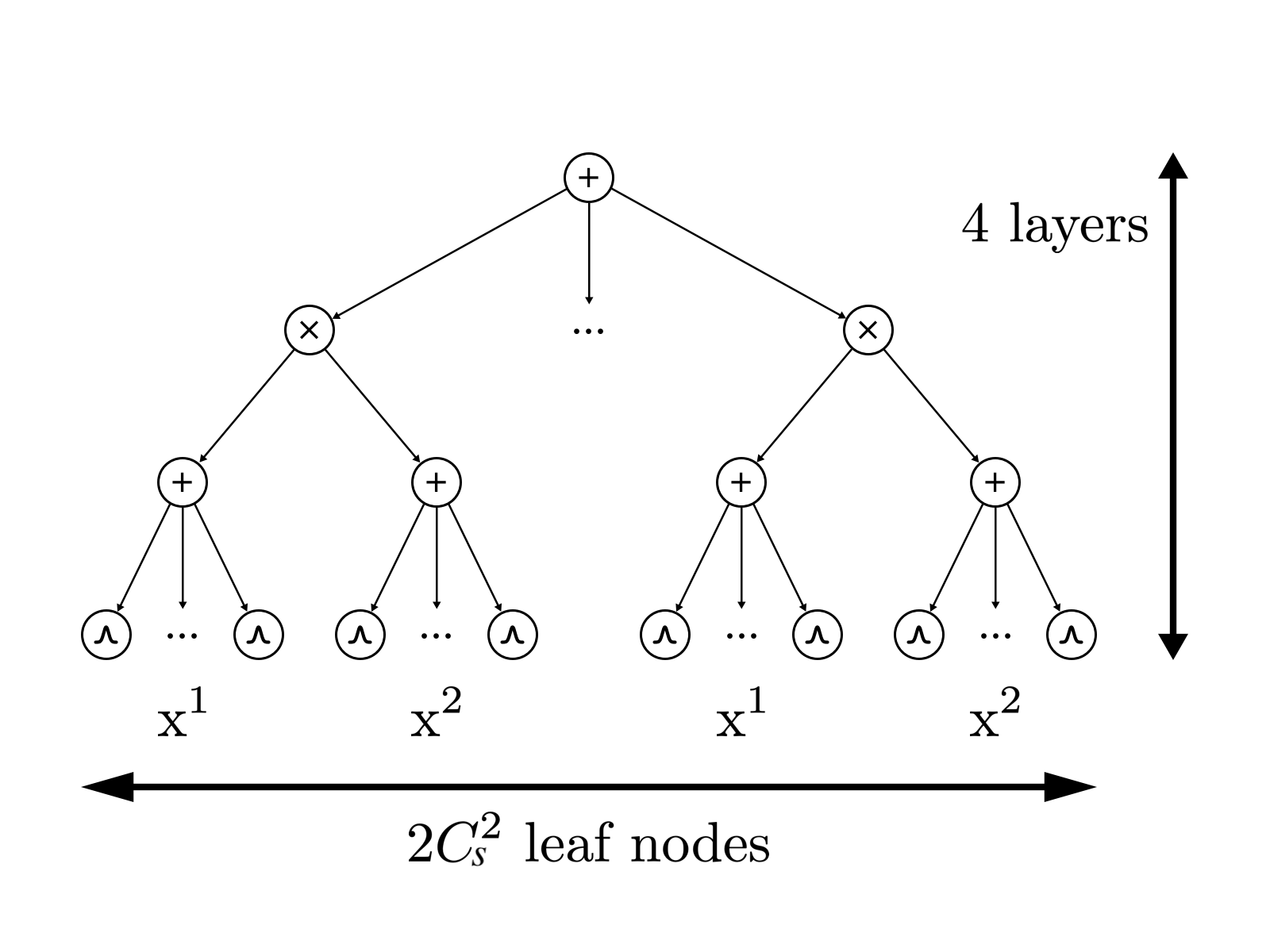}
    \end{subfigure}
    \begin{subfigure}{0.49\linewidth}
        \centering
        \includegraphics[keepaspectratio, width=\linewidth,height=10cm]{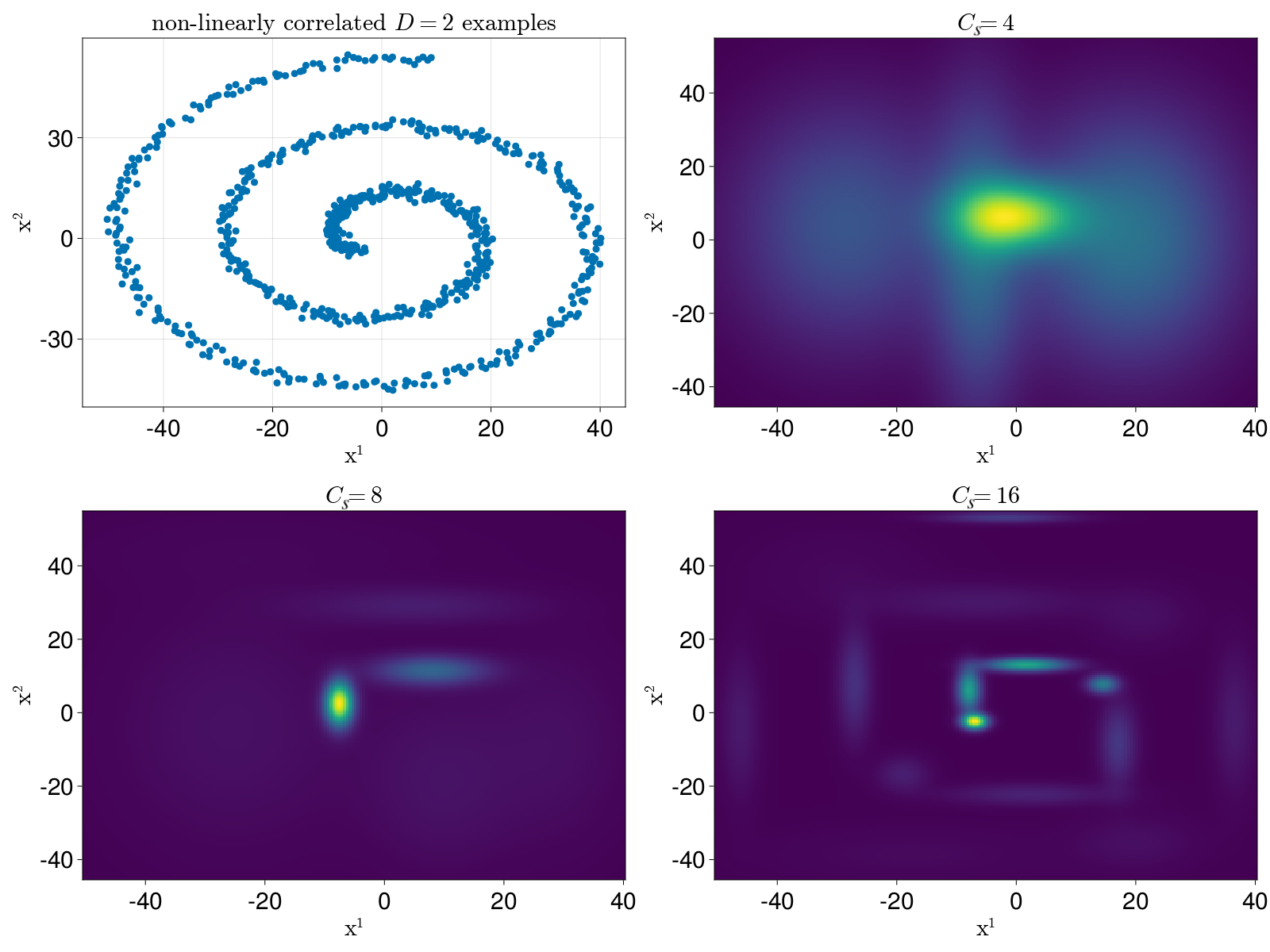}
    \end{subfigure}
    \caption{Example of the SPN for $D=2$ input data. Left: the largest computational graph with $C_s$ children for each sum node. Right: graph breadth and expressivity on non-linearly correlated examples.}
    \label{fig:SPNlimitation2D}
\end{figure}

\Cref{fig:SPNlimitation2D} left shows the possible graph structures of an SPN for $D=2$ data. 
While the height is limited to a maximum of $4$ when $C_p=2$, the breadth can expand at $2C_s^2$ by increasing $C_s$.
\Cref{fig:SPNlimitation2D} right depicts the density of an SPN modeling a two-dimensional artificial dataset with a strong correlation.
As $C_s$ increases and the SPN becomes broader, it can capture the complex distribution of spiral data.
SPNs are highly expressive even for complicated data by combining multiple one-dimensional distribution nodes. 
However, the breadth of SPNs significantly impacts this.
In order for SPNs to have high representational power under the structural constraints, $C_s$ must be increased.

\begin{table}[t]
    \caption{Complexity of the SPN}
    \label{tab:SPN_model_complexity}
    \centering
    \resizebox{\linewidth}{!}{
    \begin{tabular}{llllll}
        \toprule
            & & \multicolumn{2}{l}{Complete Tree}                                  & \multicolumn{2}{l}{Skewed Tree} \\
         & Definition & Maximum Size & Asymptotic Order & Maximum Size & Asymptotic Order \\
        \midrule
        $D$   & input dimension & - & - & - & - \\
        $C_s$ & sum outdegree & - & - & - & - \\
        $C_p$ & product outdegree & $D$ & - & $D$ & - \\
        -     & graph height               & $2\log_{C_p}D + 2$ & - & $2\frac{D-1}{C_p-1}+2$ & - \\
        -     & graph breadth              & $L$ & $\order{C_s^{\log_{C_p}D\bad{\bm+1}}}$ & $L$ & $\order{C_s^{\frac{D-1}{C_p-1}\bad{\bm+1}}}$ \\
        $V$ & \# all nodes & $S+P+L$ & $\order{C_s^{\log_{C_p}D\bad{\bm+1}}}$ & $S+P+L$ & $\order{C_s^{\frac{D-1}{C_p-1}\bad{\bm+1}}}$ \\
        $S$ & \# sum nodes & $\frac{(C_p C_s)^{\log_{C_p}D+1}-1}{C_p C_s-1}$ & $\order{C_s^{\log_{C_p}D}}$ & $\frac{C_p C_s^{\frac{D-1}{Cp-1}+1} + \qty(1-C_p)C_s}{C_s-1}$ & $\order{C_s^{\frac{D-1}{C_p-1}}}$ \\
        $P$ & \# product nodes & $\frac{C_s\cdot(C_p C_s)^{\log_{C_p}D}-C_s}{C_p C_s-1}$ & $\order{C_s^{\log_{C_p}D}}$ & $\frac{C_s^{\frac{D-1}{Cp-1}+1} -C_s}{C_s-1}$ & $\order{C_s^{\frac{D-1}{C_p-1}}}$ \\
        $L$ & \# distribution nodes & $C_s\cdot(C_p C_s)^{\log_{C_p}D}$ & $\order{C_s^{\log_{C_p}D\bad{\bm+1}}}$ & $\frac{\qty(C_pC_s-1)C_s^{\frac{D-1}{Cp-1}+1} + \qty(1-C_p)C_s^2}{C_s-1}$ & $\order{C_s^{\frac{D-1}{C_p-1}\bad{\bm+1}}}$ \\
        -     & \# induced trees & $C_s^{C_p\frac{D-1}{C_p-1}+1}$ & $\order{C_s^{\bad{\bm C_p\frac{D-1}{C_p-1} +1}}}$ & $C_s^{C_p\frac{D-1}{C_p-1}+1}$ & $\order{C_s^{\bad{\bm C_p}\frac{D-1}{C_p-1}\bad{\bm +1}}}$ \\
        \bottomrule
    \end{tabular}
    }
\end{table}

\subsection{Marginalized Posterior Distribution}
The existing studies~\citep{Vergari19,Trapp19} use Gibbs sampling that updates all random variables $\rvz_n \ (n=1,\ldots,N)$, $\rmW$, and $\rmTheta$ alternately in \Cref{equ:SPN_previous_conditional}, resulting in strong correlation between consecutive samples. 
In such cases, the mixing can be slow due to potential barriers that cannot be overcome unless multiple random variables are updated simultaneously. 
We solve this problem by marginalization.

From the dependency between variables shown in \Cref{fig:SPN_Bayes} left, we marginalize the two random variables $\rmW$ and $\rmTheta$.
The marginalized posterior distribution including only $\rmZ$ is given by
\begin{equation}\label{equ:SPNproposed_posterior}
    p\qty(\rmZ \mid \rmX, \valpha, \vgamma) \propto p\qty(\rmX \mid \rmZ, \vgamma) p\qty(\rmZ \mid \valpha).
\end{equation}
Sampling $\rmW$ and $\rmTheta$ can be omitted entirely during learning.
It reduces the number of variables that need to be sampled, thereby reducing the sample correlation and the number of iterations.
The parameters $\rmW$ and $\rmTheta$ are required during inference, so they are sampled immediately before use.
Since the learning and inference processes are usually separated in Gibbs sampling, this delayed evaluation approach works efficiently. 

In the algorithm, the sampling of $\rmW$ and $\rmTheta$ is replaced by deterministic computation of sufficient statistics, which is constantly referenced during sampling $\rmZ$. 
The problem here is that marginalization can sometimes result in complex implementation, leading to decreased performance.
We will explain the new sampling algorithm designed from the viewpoint of computational complexity.

\subsection{Top-Down Sampling Method}
The full conditional distribution of Gibbs sampling to obtain a sample from the posterior is given by
\begin{equation}\label{equ:SPNproposed_fullconditional}
    p\qty(\rvz_n \mid \rmX, \rmZ_{\setminus n}, \valpha, \vgamma).
\end{equation}

Directly generating $\rvz_n$ from \Cref{equ:SPNproposed_fullconditional} is difficult due to the dependencies of random variables.
We break it down into simple components.
The distribution can be interpreted as a product of two factors:
\begin{equation}\begin{split}\label{equ:SPN_propose_marginalposterior}
    p\qty(\rvz_n = \vc \mid \rmX, \rmZ_{\setminus n}, \valpha, \vgamma) \ \propto \ \underbrace{p\qty(\rvz_n = \vc \mid \rmZ_{\setminus n}, \valpha)}_{\textrm{network}}
    \ \cdot \ \underbrace{p\qty(\rvx_n \mid \rmX_{\setminus n}, \rvz_n = \vc, \rmZ_{\setminus n}, \vgamma)}_{\textrm{leaf}}.
\end{split}\end{equation}
The network determines the first factor, i.e., the weights of the sum nodes. 
The leaves determine the second factor, i.e., the probability distributions of the distribution nodes. 
These factors have different properties and thus require different approaches for efficient computation.
We design our algorithm to have calculations related to $S$ and $D$, where the degree of increase is relatively small from the complexity of SPNs in \Cref{tab:SPN_model_complexity}.

\subsubsection{Network Proposal}
By the conjugate prior distribution of the Multinomial and Dirichlet distributions in \Cref{equ:SPN_modeling_z}, the network factor is Dirichlet posterior predictive distributions after integrating out $\rmW$
\begin{equation}\begin{split}\label{equ:SPN_propose_networkfactor}
    p\qty(\rvz_n = \vc \mid \rmZ_{\setminus n}, \valpha) 
    &\propto \prod_{s=1}^S \qty(\mathrm{N}_{\setminus n}^{s,c^s} + \evalpha_{\evc^s})
\end{split}\end{equation}
represented by allocation counts $\mathrm{N}_{\setminus n}^{s,c} = \sum_{m=(1,\ldots,N)\setminus n} \operatorname{\delta} \qty(\ervz_m^s = c)$ and the concentration hyperparameters $\valpha$.

\Cref{equ:SPN_propose_networkfactor} can be evaluated and sampled with $\order{S}$ time.
We regard \Cref{equ:SPN_propose_networkfactor} as mixtures of allocation counts and concentration parameters for $s \in(1,\ldots,S)$
\begin{equation}\label{equ:SPN_propose_networkfactor_gen}
    \widehat{\ervz}_n^s \sim
    \begin{cases}
        \mathrm{N}_{\setminus n}^{s,c^s}  & \text{with probability $\propto \sum_{c}\mathrm{N}_{\setminus n}^{s,c}$} \\
        \evalpha_{\evc^s}                 & \text{with probability $\propto \sum_{c}\evalpha_{\evc}$}.
    \end{cases}
\end{equation}
This can be implemented with straightforward memory lookup and basic random number generation.
With probabilities proportional to $\sum_{c}\mathrm{N}_{\setminus n}^{s,c}$, sampling from $\mathrm{N}_{\setminus n}^{s,c}$ for $c\in(1,\ldots,C_s)$ is performed, which can be immediately obtained by uniformly selecting a single element from previous allocations $\rvz_{\setminus n}^s$.
With probabilities proportional to $\sum_{c}\evalpha_{\evc}$, sampling from $\evalpha_c$ is performed, which can be obtained by uniform or alias sampling. 
These steps efficiently generate candidate $\widehat{\rvz}_n = \vc$ from the network factor.

The sampling time complexity $\order{S}$ is practically important. 
The number of sum nodes $S$ is approximately $\order{C_s^{\log_{C_p}D}}$, 
and it does not increase extremely as $D$ and $C_s$ increase, unlike $\order{C_s^{\log_{C_p}D+1}}$ for the number of all nodes $V$. 
Therefore, the above steps can be executed stably in large SPNs.

To make the $\widehat{\rvz}_n = \vc$ generated by this method follow \Cref{equ:SPNproposed_fullconditional}, the rejection described next is necessary.

\subsubsection{Leaf Acceptance}

By the conjugate prior of leaf probability distribution in \Cref{equ:SPN_modeling_x}, the leaf factor, after integrating out $\rmTheta$, is a posterior predictive distribution of the leaves in the induced tree $T(\vc)$ given state $\rvz_n = \vc$:
\begin{equation}\begin{split}\label{equ:SPN_propose_leaffactor}
    p\qty(\rvx_n \mid \rmX_{\setminus n}, \rvz_n = \vc, \rmZ_{\setminus n}, \vgamma)
    & = \prod_{L_j^d\in T(\vc)} p_{L_j^d}\qty(\ervx_n^d\mid\rvx_{\setminus n}^d,\rmZ_{\setminus n},\evgamma_j^d),
\end{split}\end{equation}
where $p\qty(\rvx_n\mid\rvz_n=\vc,\rmTheta)$ is the output of $T(\vc)$, and the posterior predictive distribution of leaf $L_j^d$ is given by
\begin{equation}\begin{split}\label{equ:SPN_propose_leafpred}
    p_{L_j^d}&\qty(\ervx_n^d\mid\rvx_{\setminus n}^d,\rmZ_{\setminus n},\evgamma_j^d) = \int L_j^d\qty(x_n^d\mid\rtheta_j^d) p\qty(\rtheta_j^d\mid\rvx_{\setminus n}^d,\rmZ_{\setminus n},\evgamma_j^d)d\ervtheta_j^d.
\end{split}\end{equation}
The induced tree has one distribution node per feature dimension $d$ in the complete and decomposable SPN.
Referring to only $D$ distribution nodes is sufficient in the leaf factor, which is less costly than calculating all distribution nodes.
The problem is that the combination $\{L_j^d\}_{d=1}^D$ depends on the graph structure of the SPN.
The joint sampling for $\{L_j^d\}_{d=1}^D$ cannot be reduced to simple per-dimension calculations.
Enumerating the possible patterns of induced trees requires $\order{C_s^{\frac{C_p D-1}{C_p-1}}}$ time as shown in \Cref{tab:SPN_model_complexity}, exhibiting a steep growth as $D$ and $C_s$ increase.
Generating candidates $\widehat{\rvz}_n = \vc$ from \Cref{equ:SPN_propose_leaffactor} is computationally prohibited.

Instead, we use the leaf factor to accept candidates. 
To make the candidate $\widehat{\rvz}_n = \vc$ generated by the network factor follow the marginalized posterior distribution, the Metropolis--Hastings acceptance probability of the move from $\rvz_n=\bm{b}$ to $\vc$ is $\min\qty(1,A\qty(\bm{b}\to\vc))$ where
\begin{equation}\begin{split}\label{equ:SPN_propose_acceptance}
    A\qty(\bm{b}\to\vc)
    &= \frac{
        \prod_{L_j^d\in T(\vc)} p_{L_j^d}\qty(\ervx_n^d\mid\rvx_{\setminus n}^d,\rmZ_{\setminus n},\evgamma_j^d)
    }{
        \prod_{L_j^d\in T(\bm{b})} p_{L_j^d}\qty(\ervx_n^d\mid\rvx_{\setminus n}^d,\rmZ_{\setminus n},\evgamma_j^d)
    }.
\end{split}\end{equation}
The network factors in the numerator and denominator cancel out each other and do not need to be evaluated. 

By computing the sufficient statistics of the distribution nodes beforehand, \Cref{equ:SPN_propose_leaffactor} can be evaluated in $\order{D}$, independent of the number of datapoints $N$.
Since $p_{L_j^d}$ is a one-dimensional probability distribution, the sufficient statistics can be easily updated in constant time for each sample.
As shown in \Cref{tab:SPN_model_complexity}, $D$ is tiny compared to other dimensionalities, so the acceptance probability can be efficiently obtained.
Furthermore, the evaluation of predictive distributions can be omitted for the dimension $d$ where the candidate component does not change ($b^d=c^d$).
It accelerates Gibbs sampling even when the samples are correlated.

\begin{table}[t]
    \caption{Time complexity of the sampling algorithms}
    \label{tab:SPN_propose_samplingcomplexity}
    \centering
    \begin{tabular}{cll}
        \toprule
                                        & Top-down                      & Bottom-up \citep{Vergari19} \\
        \midrule
        Gibbs sampling & $\order{C_s^{\log_{C_p}D}}$     & $\order{C_s^{\log_{C_p}D \bad{\bm+1}}}$ \\
        inference with pre-processing & $\order{C_s^{\log_{C_p}D \bad{\bm+1}}}$ & $\order{C_s^{\log_{C_p}D \bad{\bm+1}}}$ \\
        \bottomrule
    \end{tabular}
\end{table}

\subsubsection{Algorithm and Complexity}

The efficient implementation of this algorithm involves traversing the computational graph top-down while referring to the candidate state $\widehat{\rvz}_n = \vc$ generated by the network factor, as shown in \Cref{fig:SPN_Bayes} right. 
Ignoring the nodes outside the induced tree, only the distribution nodes required in \Cref{equ:SPN_propose_acceptance} are evaluated.
Compared to the existing method traversing the entire graph bottom-up in \Cref{fig:SPN_Bayes} center, the proposed method can be more efficiently executed by accessing a subgraph consisting of a limited number of nodes.
Intuitively, the top-down method is more efficient for wider graphs.
\Cref{tab:SPN_propose_samplingcomplexity} compares the time complexity of the algorithms.
The entire algorithm of the top-down method is shown as \Cref{sec:propose_algorithm}.

The parameters $\rmW$ and $\rmTheta$ marginalized by the proposed method are not updated during Gibbs sampling. 
These values need to be updated to the latest ones in inference time. 
This pre-processing is identical to what is done during each iteration of Gibbs sampling in the bottom-up algorithm (\Cref{equ:SPN_previous_conditional}) and is typically performed at checkpoints during training or upon completion of training.
Since these are not significant in terms of time complexity, the inference time complexity of the proposed method does not change as shown in \Cref{tab:SPN_propose_samplingcomplexity}.

\subsection{Empirical Bayesian Hyperparameter Tuning}\label{sec:SPN_propose_hyperparameter}

Another challenge in large-scale SPNs is hyperparameter optimization, an essential task in Bayesian learning. 
The hyperparameters of SPNs include $C_s$, $C_p$, $\valpha$, and $\vgamma$. 
In particular, the appropriate $\vgamma_j^d$ induces diversity in each distribution node $L_j^d$ and helps SPNs choose suitable component distributions.
As shown in \Cref{tab:SPN_model_complexity}, there are asymptotically $\order{C_s^{\log_{C_p}D + 1}}$ distribution nodes and $\vgamma$ is proportional to them.
The number of hyperparameters can easily exceed hundreds to thousands in a real-world dataset. 
Applying currently common hyperparameter optimization methods such as tree-structured Parzen estimator~(TPE)~\citep{Bergstra:2011aa} is not advised in terms of the number of trials.
We must consider tuning proxy parameters instead of directly tuning $\vgamma$.

In the empirical Bayesian approach, hyperparameters are obtained by maximizing the marginal likelihood. 
Although calculating the marginal likelihood $p\qty(\rmX \mid \valpha, \vgamma)$ of an SPN is difficult, maximizing it with respect to a hyperparameter $\vgamma_j^d$ can be reduced to maximizing the mixture of leaf marginal likelihoods $p_{L_j^d}\qty(\cdot \mid \vgamma_j^d)$:
\begin{equation}\begin{split}\label{equ:SPN_propose_exactmarginal}
    \argmax_{\vgamma_j^d} p\qty(\rmX \mid \valpha, \vgamma)
    &= \argmax_{\vgamma_j^d} \sum_{\rmX_c \subseteq \rmX} \omega_c p_{L_j^d}\qty(\rvx^d_c \mid \vgamma_j^d).
\end{split}\end{equation}
This is derived from the fact that SPNs can be considered a mixture of induced trees \citep{zhaoUnifiedApproachLearning2016} and induced trees are factorized leaf models for each feature dimension (\Cref{sec:background_latent}).
The summation $\sum_{\rmX_c \subseteq \rmX}$ is taken over all possible subset $\rmX_c$ of dataset $\rmX$.
When there are $N$ datapoints in the dataset, the summation size is $2^N$.
The coefficient $\omega_c$ is complicated and depends on the other hyperparameters $\vgamma\setminus\evgamma_j^d$. 
Whereas the exact evaluation of \Cref{equ:SPN_propose_exactmarginal} is computationally impossible, it gives an essential insight that the marginal likelihood of distribution nodes over subset data gives the empirical Bayes estimate.

We consider approximating \Cref{equ:SPN_propose_exactmarginal} by a significant term of the leaf marginal likelihood with specific subset data.
Our goal is not to identify the optimal subset $\rmX_c$ directly but to find the subsampling ratio $r^d\in(0,1]$ that $\rmX_c$ should contain from the dataset $\rmX$ for each feature dimension $d$. 
The empirical Bayes estimate of hyperparameter $\evgamma_j^d$ is approximated with subset data $\rvx_c^d = \operatorname{subsample}\qty(\rvx^d, r^d)$ by
\begin{equation}\begin{split}
    \widehat{\evgamma}_j^d &= \argmax_{\evgamma_j^d} p_{L_j^d}\qty(\rvx^d_c \mid \evgamma_j^d).
\end{split}\end{equation}
By tuning $r^d$ using hyperparameter optimization methods, distribution nodes with appropriate hyperparameters are expected to become the main component of the leaf mixture. 
When $r^d=1$, it is empirical Bayes over the full dataset, and all the distribution nodes $L^d$ have the same hyperparameters for $d$. 
As $r^d$ approaches $0$, the intersection of the subset data becomes smaller, resulting in diverse distribution nodes. 
This approach significantly reduces the number of optimized hyperparameters from the order of distribution nodes $\order{C_s^{\log_{C_p}D+1}}$ to $\order{D}$, while maintaining the flexibility of the prior distributions. 

It is easy to obtain the empirical Bayes solution for leaf hyperparameters numerically because distribution nodes have one-dimensional probability distributions.
Closed-form solutions can also be obtained for some parameters. 
\Cref{sec:distributionnodes} summarizes the conjugate prior, posterior, posterior predictive, and closed-form empirical Bayesian hyperparameters for typical distribution nodes.

\subsection{Contrast with Prior Work} \label{sec:SPN_discussion}

\citet{zhaoCollapsedVariationalInference2016} employed collapsed variational inference to approximate the posterior distribution. 
Their algorithm is optimization-based and significantly different from our method based on posterior sampling. 
Also, it integrated out $\rmZ$, which is a unique approach different from many other variational inferences. 
It is also distinct from our posterior distribution marginalizing $\rmW$ and $\rmTheta$. 

The proposed method only discusses Bayesian parameter learning and does not mention structural learning. 
In the experiments in \Cref{sec:SPN_experiment}, we use a heterogeneous leaf mixture similar to \citet{Vergari19}, and an appropriate probability distribution is selected by weighting from multiple types of distribution nodes.
It is possible to perform Bayesian structural learning similar to \citep{Trapp19}, but pre-processing is required for the marginalized parameters $\rmTheta$ for a conditional probability in structural inference.
The speed benefits of our method may be compromised depending on the frequency of structural changes.

\citet{Vergari19,Trapp19} assume that all distribution nodes have the same hyperparameters for each feature dimension and distribution type.
This simple setting always assumes the same prior distribution, so it does not induce diversity in distribution nodes like our empirical Bayesian approach on subset data. 
When the hyperparameters are selected to maximize the marginal likelihood, the results are expected to be similar to the proposed method with subsampling proportion $r^d=1 \ \ \forall d$.

\section{Experiments}\label{sec:SPN_experiment}
In this section, we compare the empirical performance of top-down and bottom-up sampling methods. 
We used a total of $24$ datasets, where $18$ datasets were from the UCI repository\,\citep{UCIrepo} and OpenML\,\citep{OpenML} and $6$ datasets were from the previous studies (Abalone, Breast, Crx, Dermatology, German, Wine). 
\Cref{sec:dataset_details} shows the dimensions of each dataset.
The datasets were divided into 8:1:1 for training, validation, and testing.
To prevent irregular influence on the results, $k$-nearest neighbor based outlier detection was performed as pre-processing.
We fixed $C_p=2$ and investigated the effect of different $C_s$.
Each configuration was tested $10$ times with different random seeds and the mean and standard deviation were plotted.
The leaf distribution node consisted of a heterogeneous mixture of one-dimensional probability distributions similar to \citet{Vergari19} (cf. \Cref{sec:distributionnodes}).
The hyperparameters of the leaf distributions were obtained by the empirical Bayes approach described in \Cref{sec:SPN_propose_hyperparameter} optimized by conducting $100$ trials of TPE. 
The code was written in Julia and iterated as much as possible within 12 hours (including a 6-hour burn-in period) on an Intel Xeon Bronze 3204 CPU machine.

\subsection{Computational Efficiency}
We first measured the elapsed time required for each sampling method to perform one iteration of Gibbs sampling and confirmed how much acceleration was achieved. 
\Cref{tab:SPN_exp_itertime} shows the elapsed time required for each sampling method to perform one iteration of Gibbs sampling.
The results consistently show that the top-down method is superior and generally several orders of magnitude shorter in execution time.
The bottom-up method calculates outputs at all nodes and propagates them using the \texttt{logsumexp} algorithm, whereas the top-down method only identifies the memory addresses of induced leaves. Therefore, the actual computation time is significantly faster than the theoretical time complexity suggests.
In particular, the difference is significant as $C_s$ increases.
These results show that the top-down method is tens to more than one hundred times faster, supporting the efficiency of our method.

\begin{table*}[h]
    \caption{Elapsed time [s] per Gibbs iteration (mean $\pm$ std)}
    \label{tab:SPN_exp_itertime}
    \centering
    \resizebox{0.8\linewidth}{!}{
    \begin{tabular}{lrrrrrr}
        \toprule
             & \multicolumn{3}{l}{$C_s=2$}                                  & \multicolumn{3}{l}{$C_s=4$} \\
        Dataset & Top-down                    & Bottom-up       & \textit{Speedup} & Top-down           & Bottom-up       & \textit{Speedup} \\
        \midrule
        Abalone            & {\boldmath $0.076\pm0.004$} & $1.574\pm0.043$ & $\times21$ & {\boldmath $0.367\pm0.019$} & $17.38\pm0.699$ & $\times47$ \\
        Ailerons           & {\boldmath $2.938\pm0.121$} & $96.28\pm2.631$ & $\times33$ & {\boldmath $181.2\pm28.06$} & $3282\pm105.9$  & $\times18$ \\
        Airfoil Self-Noise & {\boldmath $0.010\pm0.001$} & $0.303\pm0.024$ & $\times30$ & {\boldmath $0.021\pm0.001$} & $2.047\pm0.143$ & $\times97$ \\
        Breast             & {\boldmath $0.005\pm0.000$} & $0.170\pm0.016$ & $\times34$ & {\boldmath $0.014\pm0.001$} & $1.938\pm0.079$ & $\times138$ \\
        Computer Hardware  & {\boldmath $0.003\pm0.000$} & $0.098\pm0.011$ & $\times33$ & {\boldmath $0.007\pm0.000$} & $0.943\pm0.078$ & $\times135$ \\
        cpu\_act            & {\boldmath $0.525\pm0.016$} & $16.80\pm0.771$ & $\times32$ & {\boldmath $11.72\pm2.503$} & $588.0\pm15.46$ & $\times50$ \\
        cpu\_small          & {\boldmath $0.204\pm0.005$} & $5.592\pm0.206$ & $\times27$ & {\boldmath $2.059\pm0.119$} & $99.17\pm3.382$ & $\times48$ \\
        Crx                & {\boldmath $0.023\pm0.001$} & $0.722\pm0.013$ & $\times31$ & {\boldmath $0.232\pm0.006$} & $14.21\pm0.458$ & $\times61$ \\
        \bottomrule
    \end{tabular}
    }
\end{table*}

\begin{table*}[h]
    \begin{minipage}{.5\linewidth}
        \caption{Effective sample size per 50 samples}
        \label{tab:SPN_exp_effectivesamplesize}
      \centering
      \resizebox{0.97\linewidth}{!}{
      \begin{tabular}{lrrrr}
        \toprule
                                        & \multicolumn{2}{l}{$C_s=2$}                 & \multicolumn{2}{l}{$C_s=4$} \\
        Dataset     & Top-down & Bottom-up                        & Top-down & Bottom-up \\
        \midrule
        Dermatology                     & $15.6\pm0.3$ & {\boldmath $20.6\pm0.1$} & $19.1\pm0.1$ & {\boldmath $20.1\pm0.0$} \\
        elevators                       & $15.1\pm0.2$ & {\boldmath $18.0\pm0.3$} & {\boldmath $17.0\pm0.1$} & - \\
        Forest Fires                    & $18.9\pm0.5$ & {\boldmath $20.6\pm0.2$} & $18.6\pm0.1$ & {\boldmath $20.5\pm0.0$} \\
        German                          & $17.3\pm0.3$ & {\boldmath $20.6\pm0.1$} & $18.9\pm0.1$ & {\boldmath $19.4\pm0.0$} \\
        Housing                         & $17.3\pm1.2$ & {\boldmath $20.6\pm0.1$} & $17.4\pm0.2$ & {\boldmath $20.5\pm0.0$} \\
        Hybrid Price                    & $20.4\pm0.7$ & {\boldmath $20.5\pm1.0$} & $20.5\pm0.4$ & {\boldmath $20.6\pm0.4$} \\
        kin8nm                          & $18.2\pm2.1$ & {\boldmath $20.8\pm0.3$} & $16.8\pm0.1$ & {\boldmath $19.5\pm0.2$} \\
        LPGA2008                       & {\boldmath $20.7\pm0.6$} & $20.5\pm0.2$ & {\boldmath $20.5\pm0.2$} & {\boldmath $20.5\pm0.1$} \\
        \bottomrule
    \end{tabular}
      }
    \end{minipage}
    \begin{minipage}{.5\linewidth}
      \caption{Log-likelihood}
    \label{tab:SPN_exp_loglikelihood}
    \centering
    \resizebox{\linewidth}{!}{
    \begin{tabular}{lrrrr}
        \toprule
                        & \multicolumn{2}{l}{$C_s=2$}                & \multicolumn{2}{l}{$C_s=4$} \\
        Dataset         & Top-down & Bottom-up                       & Top-down & Bottom-up \\
        \midrule
        LPGA2009       & {\boldmath $-28.3\pm0.4$} & $-29.6\pm0.2$  & {\boldmath $-27.4\pm0.4$} & $-29.7\pm0.2$ \\
        P. motor         & {\boldmath $43.9\pm0.7$} & $41.9\pm0.3$    & {\boldmath $43.5\pm0.6$} & $42.1\pm0.1$ \\
        P. total         & {\boldmath $43.4\pm0.6$} & $41.7\pm0.1$    & {\boldmath $42.8\pm0.8$} & $41.5\pm0.1$ \\
        Vote            & {\boldmath $-49.6\pm0.1$} & $-49.8\pm0.0$  & {\boldmath $-49.8\pm0.1$} & $-49.9\pm0.1$ \\
        W. Red        & {\boldmath $-3.6\pm0.3$} & $-4.4\pm0.4$    & {\boldmath $-2.7\pm0.3$} & $-3.0\pm0.1$ \\
        W. White      & {\boldmath $-3.0\pm0.6$} & $-3.6\pm0.6$    & {\boldmath $-2.6\pm0.1$} & {\boldmath $-2.6\pm0.1$} \\
        Wine            & {\boldmath $-19.9\pm0.6$} & $-22.4\pm0.5$  & {\boldmath $-18.4\pm0.4$} & $-20.6\pm0.1$ \\
        Yacht           & {\boldmath $2.2\pm1.8$} & $1.0\pm1.9$      & {\boldmath $8.3\pm0.6$} & $5.0\pm0.2$ \\
        \bottomrule
    \end{tabular}
    }
    \end{minipage} 
\end{table*}

\subsection{Sample Correlation}\label{sec:SPN_exp_samplequality}
Samples generated by Gibbs sampling are not independent but rather long sequences of random vectors that exhibit correlation over time. 
Samples with correlation are redundant and do not provide information about the underlying distribution efficiently, causing underestimation of errors in Bayesian inference.
In this experiment, we investigate the quality of the samples obtained by the two sampling methods by comparing the effective sample size.

\Cref{tab:SPN_exp_effectivesamplesize} shows the effective sample size of $50$ thinned subsamples at equal intervals.
Since the bottom-up method could not generate $50$ samples within the time limit for some datasets, the results are not shown for the cases.
The values are large enough for many configurations, indicating that both sampling methods provide sufficiently independent samples within a realistic time range.
The bottom-up method is slightly superior in many cases, but the difference is not extreme overall.
Due to the marginalization and rejection, the result of the top-down method remained within a comparable range to that of the bottom-up method.
The results are stable against the increase in $C_s$.

\subsection{Overall Performance}
Finally, we evaluated the overall predictive performance of the top-down sampling method, which reveals the trade-off between fast iteration speed and sample correlation.

\begin{figure*}[t]
    \centering
    \includegraphics[keepaspectratio, width=0.33\linewidth]{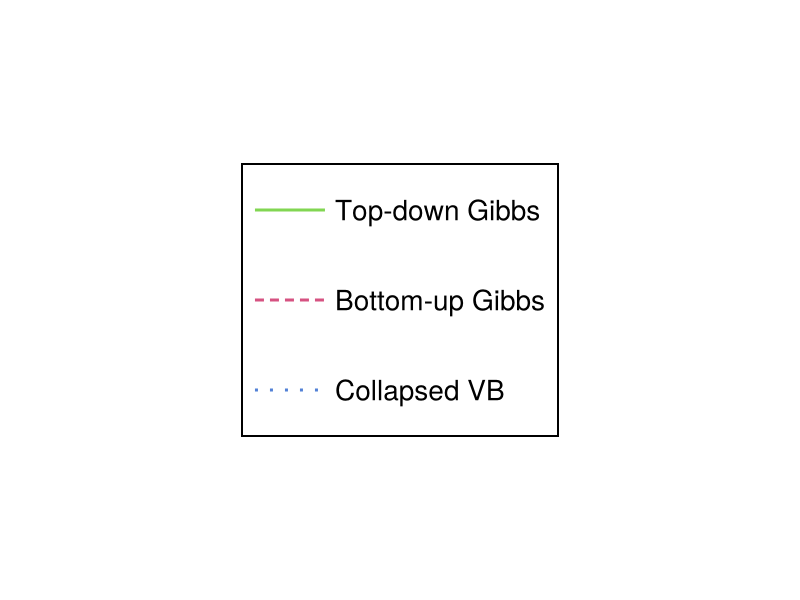}
    \includegraphics[keepaspectratio, width=0.33\linewidth]{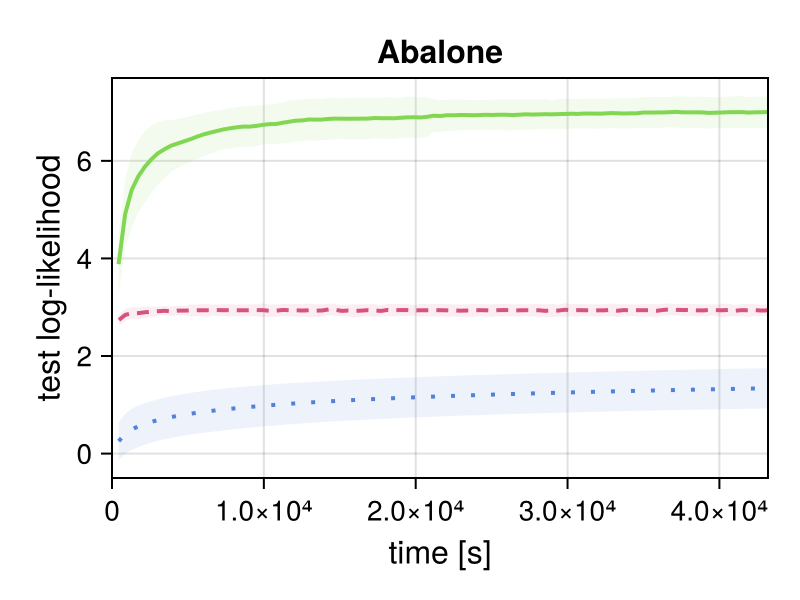}
    \includegraphics[keepaspectratio, width=0.33\linewidth]{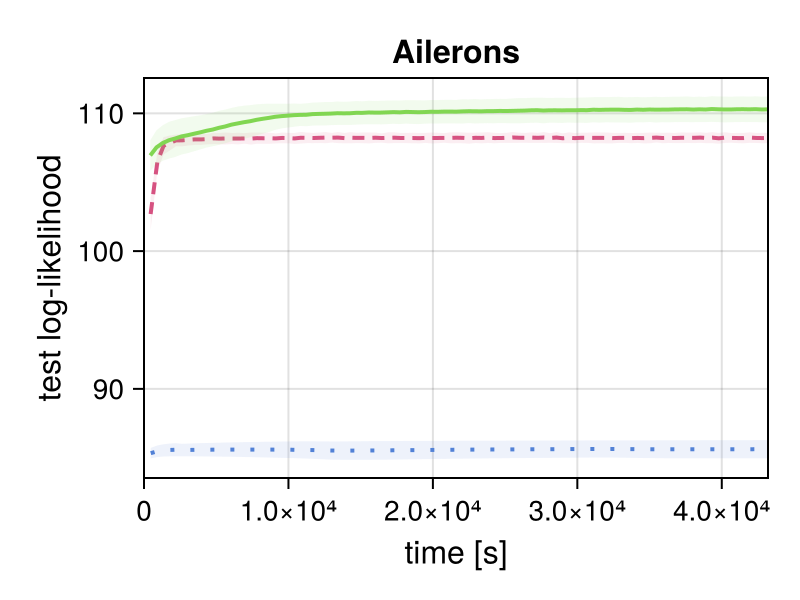}
    \\
    \includegraphics[keepaspectratio, width=0.33\linewidth]{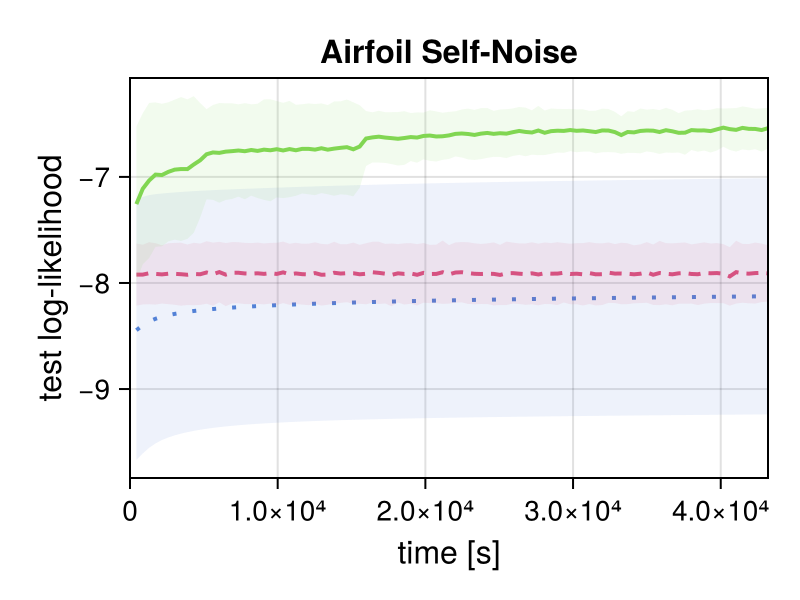}
    \includegraphics[keepaspectratio, width=0.33\linewidth]{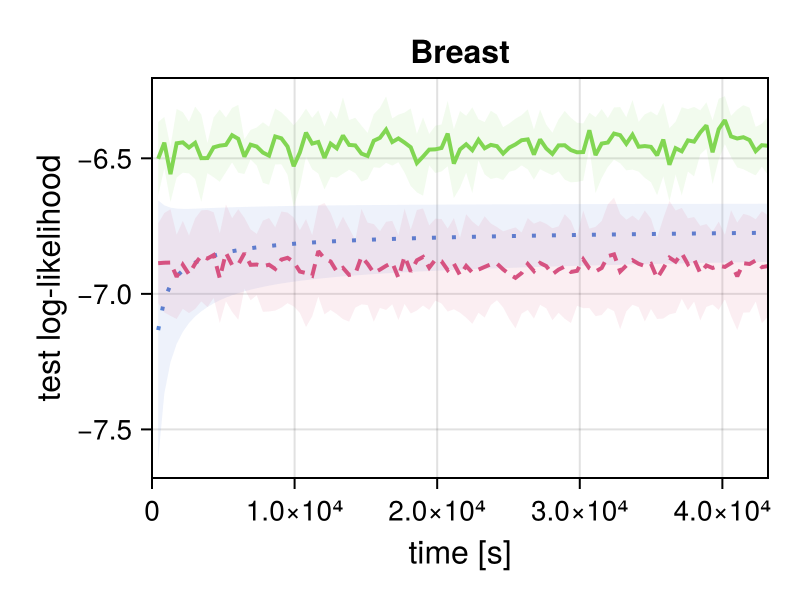}
    \includegraphics[keepaspectratio, width=0.33\linewidth]{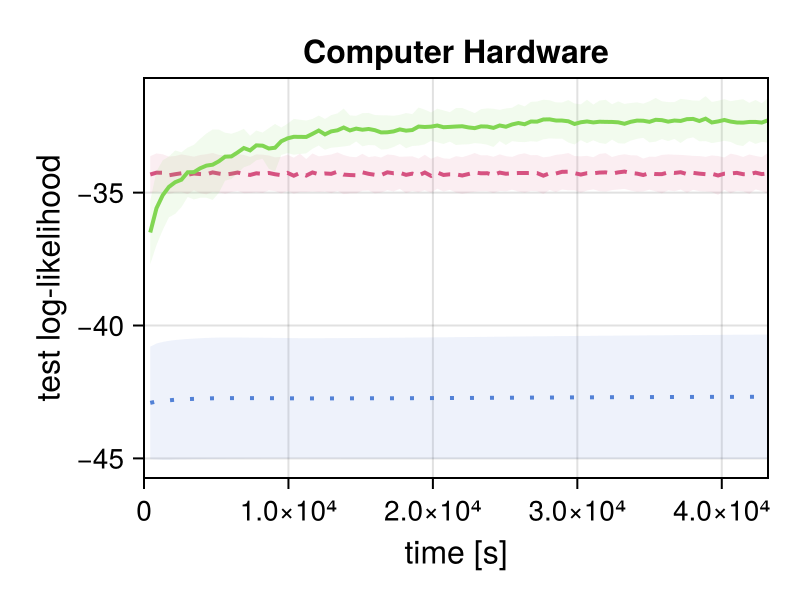}
    \caption{Temporal evolution in predictive performance for each dataset. The average test-set log-likelihood over $10$ trials is shown with the lines, and the standard deviation is indicated by the shaded regions.}
    \label{fig:SPN_exp_llh_feature}
\end{figure*}

\Cref{tab:SPN_exp_loglikelihood} compares the log-likelihood on the test set for different $C_s$, showing the results after the burn-in period.
Also, \Cref{fig:SPN_exp_llh_feature} illustrates the temporal evolution of predictive performance, showing the differences between the methods by optimizing $C_s$.
For practical interest, this experiment also compares the results with those of the collapsed variational Bayes\,\citep{zhaoCollapsedVariationalInference2016} based on optimization by the gradient descent method.
The top-down method is almost consistently superior to the others, achieving the same or higher likelihood in most configurations.
These results suggest that the algorithm speedup outweighs the impact of sample correlation.
While collapsed VB is memory efficient because it does not need to store latent variables for each data point, it requires two full network traversals for output propagation and gradient calculation, resulting in a time complexity of the same order as the bottom-up method.
For a more comprehensive set of results, please refer to \Cref{sec:all_experimental_results}.

From the above experiments, we conclude that
\begin{enumerate*}
    \item the top-down method is tens to more than one hundred times faster than the bottom-up method,
    \item the sample correlation is sufficiently small for both methods, and
    \item as a result, the top-down method can achieve higher predictive performance than the bottom-up method in many cases.
\end{enumerate*}

\clearpage

\section{Conclusion} \label{sec:SPN_conclusion}
Prior work has interpreted SPNs as latent variable models and introduced Gibbs sampling as a Bayesian learning method.
This has made it possible to automatically discriminate different distribution types for data and learn network structures in a manner consistent with the Bayesian framework.
However, the bottom-up posterior sampling approach based on the entire graph evaluation had a computational difficulty.
The shape of the computational graph was a bottleneck due to the structural constraints.

This study aimed to accomplish a fast Bayesian learning method for SPNs.
First, we investigated the complexity of SPNs when the outdegrees of the sum and product nodes are given and discussed the graph shape of SPN with high representational power.
We also derived the new full conditional probability that marginalizes multiple variables to improve sample mixing.
For the complexity and the marginalized posterior distribution, we proposed the top-down sampling algorithm based on the carefully designed proposal and rejection steps of the Metropolis--Hastings.
Our optimization efforts resulted in a time complexity reduction from $\order{C_s^{\log_{C_p}D + 1}}$ down to $\order{C_s^{\log_{C_p}D}}$.
In numerical experiments on more than $20$ datasets, we demonstrated a speedup of tens to more than one hundred times and improved predictive performance.

\clearpage

\bibliographystyle{unsrtnat}
\bibliography{ref}

\clearpage

\appendix

\section{Notation}

\begin{table}[h]
    \caption{Notation}
    \label{tab:notation}
    \centering
    \begin{tabular}{ccc}
      \toprule
      Symbol & Definition & Dimension\\
      \midrule
      $\rvx_n$ & $n$-th data, $n\in\qty(1,\ldots,N)$ & $D$ \\
      $\rvz_n$ & categorical assignments of $\vx_n$ & $S$ \\
      $s$ & sum node, $s\in\qty(1,\ldots,S)$ & $1$ \\
      $\rvw^s$ & weight vector of $s$ & $C$ \\
      $\alpha$ & hyperparameter of $\rvw^s$ & $1$ \\
      $L_j^d$ & $j$-th leaf distribution of dimension $d$ & $1$ \\
      $\ervtheta_j^d$ & parameters of $L_j^d$ & $1$ \\
      $\gamma_j^d$ & hyperparameters of $\ervtheta_j^d$ & $1$ \\
      $T(\rvz)$ & induced tree by $\rvz$ & - \\
    \bottomrule
  \end{tabular}
\end{table}

\section{Algorithm}\label{sec:propose_algorithm}

\begin{algorithm}[h]
    \caption{Bayesian learning by top-down sampling}\label{alg:SPN_propose_learning}
    \begin{algorithmic}
    \Require dataset $\rmX$, hyperparameters $\valpha$ and $\vgamma$, the number of children $C_s$ and $C_p$
    \State initialize $\operatorname{SPN}\qty(\valpha,\vgamma,C_s,C_p)$ and sample store $\rmZ \text{s}$
    \For{Gibbs iteration $i$}
      \For{data index $n \in \qty(1,\dots,N)$}
        \State generate candidate $\vc \sim \cdot \mid \rmZ_{\setminus n}, \valpha$ \Comment{\Cref{equ:SPN_propose_networkfactor_gen}}
        \If{accept $A\qty(\bm{b}\to\vc) > rand()$} \Comment{\Cref{equ:SPN_propose_acceptance}}
            \State update assignment $\rvz_n \gets \vc$
            \State update assign counts $\mathrm{N}_{\setminus n}^{s,c^s}$
            \State update leaf sufficient statistics $p_{L_j^d}$
        \EndIf
      \EndFor
      \If{$i \notin \textrm{burn-in}$}
        \State add sample $\rmZ \text{s} \gets \rmZ \text{s} \cup \rmZ$
      \EndIf
    \EndFor
    \State \textbf{return} $\rmZ \text{s}$
    \end{algorithmic}
\end{algorithm}

\begin{algorithm}[h]
    \caption{Pre-processing for inference}\label{alg:SPN_propose_inference}
    \begin{algorithmic}
    \Require assignments $\rmZ$, dataset $\rmX$, hyperparameters $\valpha$ and $\vgamma$
    \State sample sum node weights $\rmW \sim \cdot \mid \rmZ$ \Comment{\Cref{equ:SPN_previous_conditional}}
    \State sample leaf parameters $\rmTheta \sim \cdot \mid \rmX, \rmZ$
    \State \textbf{return} $\rmW$ and $\rmTheta$
    \end{algorithmic}
\end{algorithm}

\clearpage
\section{Distribution Nodes}\label{sec:distributionnodes}

\begin{table}[h]
    \caption{Distrbution nodes}
    \label{tab:SPN_distributionnodes}
    \centering
    \resizebox{\linewidth}{!}{
    \begin{tabularx}{0.9\textwidth}{c@{\hspace{0.15\linewidth}}c@{\hspace{0.1\linewidth}}c}
        \toprule
                    & Likelihood                                      & Conjugate Prior \\
        \midrule
        Exponential & $\mathrm{Exp}\qty(x\mid\lambda)$ & $\mathrm{Gamma}(\lambda\mid\alpha,\beta)$ \\
        Gaussian    & $\mathrm{N}\qty(x \mid \mu, \tau^{-1})$ & $\mathrm{N}\qty(\mu \mid \mu_0, \qty(\rho_0\tau)^{-1}) \mathrm{Gamma}(\tau \mid a_0, b_0)$ \\
        Poisson     & $\mathrm{Poisson}\qty(x\mid\lambda)$ & $\mathrm{Gamma}(\lambda\mid\alpha,\beta)$ \\
        Multinomial & $\mathrm{Multi}\qty(x\mid\vpi)$ & $\mathrm{Dir}\qty(\vpi\mid\valpha)$ \\
    \end{tabularx}
    }
    \resizebox{\linewidth}{!}{
    \begin{tabularx}{0.9\textwidth}{ccc}
        \toprule
                    & Posterior & Predictive \\
        \midrule
        Exponential & $\mathrm{Gamma}\qty(\lambda\mid\alpha_N,\beta_N)$ & $\mathrm{Lomax}\qty(x_{N+1}\mid\alpha_N,\beta_N)$ \\
        Gaussian    & $\mathrm{N}\qty(\mu \mid \mu_N, \qty(\rho_N\tau)^{-1}) \mathrm{Gamma}(\tau \mid a_N, b_N)$ & $\mathrm{St}\qty(x_{N+1} \mid \mu_N, 2a_N, 2\qty(1+\frac{1}{\rho_N})b_N)$ \\
        Poisson     & $\mathrm{Gamma}\qty(\lambda\mid\alpha_N,\beta_N)$ & $\mathrm{NegBin}\qty(x_{N+1}\mid\alpha_N,\frac{\beta_N}{\beta_N+1})$ \\
        Multinomial & $\mathrm{Dir}\qty(\vpi\mid\valpha_N)$ & $\mathrm{Multi}\qty(x_{N+1}\mid\valpha_N)$\\
    \end{tabularx}
    }
    \resizebox{\linewidth}{!}{
    \begin{tabularx}{0.9\textwidth}{c>{\centering\arraybackslash}X}
        \toprule
                    & Closed-form hyperparameters \\
        \midrule
        Exponential & $\beta = \mathrm{mean}\qty(\rvx_c^d) \alpha$ \\
        Gaussian    & $\mu_0=\mathrm{mean}\qty(\rvx_c^d)$, $b_0 = \operatorname{Var}(\rvx_c^d)a_0$ \\
        Poisson     & $\beta = \mathrm{mean}\qty(\rvx_c^d)^{-1} \alpha$ \\
        Multinomial & - \\
        \bottomrule
    \end{tabularx}
    }
\end{table}

\clearpage
\section{Dataset Details}\label{sec:dataset_details}

\begin{table}[h]
    \caption{The dimensions of datasets and corresponding SPNs}
    \label{tab:SPN_experiment_datasets}
    \centering
    \begin{tabular}{lrrrrrr}
        \toprule
        \multicolumn{3}{l}{Dataset}    & \multicolumn{2}{l}{SPN ($C_s=2$)} & \multicolumn{2}{l}{SPN ($C_s=4$)} \\
                                        & $D$ & $N$ & $S$ & $V$ & $S$ & $V$ \\
        \midrule
        Abalone                         & $9$ & $3,850$ & $117$ & $351$ & $1,097$ & $5,485$ \\
        Ailerons                        & $41$ & $12,614$ & $2,517$ & $7,551$ & $111,177$ & $555,885$ \\
        Airfoil Self-Noise              & $6$ & $1,398$ & $53$ & $159$ & $329$ & $1,645$ \\
        Breast                          & $10$ & $277$ & $149$ & $447$ & $1,609$ & $8,045$ \\
        Computer Hardware               & $9$ & $192$ & $117$ & $351$ & $1,097$ & $5,485$ \\
        cpu\_act                         & $22$ & $7,486$ & $725$ & $2,175$ & $16,969$ & $84,845$ \\
        cpu\_small                       & $13$ & $7,486$ & $245$ & $735$ & $3,145$ & $15,725$ \\
        Crx                             & $16$ & $653$ & $341$ & $1,023$ & $4,681$ & $23,405$ \\
        Dermatology                     & $35$ & $358$ & $1,749$ & $5,247$ & $62,025$ & $310,125$\\
        elevators                       & $19$ & $15,248$ & $533$ & $1,599$ & $10,825$ & $54,125$ \\
        Forest Fires                    & $13$ & $474$ & $245$ & $735$ & $3,145$ & $15,725$ \\
        German                          & $21$ & $1,000$ & $661$ & $1,983$ & $14,921$ & $74,605$ \\
        Housing                         & $14$ & $461$ & $277$ & $831$ & $3,657$ & $18,285$ \\
        Hybrid Price                    & $4$ & $146$ & $21$ & $63$ & $73$ & $365$ \\
        kin8nm                          & $9$ & $7,688$ & $117$ & $351$ & $1,097$ & $5,485$ \\
        LPGA2008                       & $7$ & $145$ & $69$ & $207$ & $457$ & $2,285$ \\
        LPGA2009                       & $12$ & $135$ & $213$ & $639$ & $2,633$ & $13,165$ \\
        Parkinsons Telemonitoring (motor) & $17$ & $5,423$ & $405$ & $1,215$ & $6,729$ & $33,645$ \\
        Parkinsons Telemonitoring (total) & $17$ & $5,412$ & $405$ & $1,215$ & $6,729$ & $33,645$ \\
        Vote for Clinton                & $10$ & $2,470$ & $149$ & $447$ & $1,609$ & $8,045$ \\
        Wine Quality Red                & $12$ & $1,469$ & $213$ & $639$ & $2,633$ & $13,165$ \\
        Wine Quality White              & $12$ & $4,495$ & $213$ & $639$ & $2,633$ & $13,165$ \\
        Wine                            & $14$ & $178$ & $277$ & $831$ & $3,657$ & $18,285$ \\
        Yacht Hydrodynamics             & $7$ & $288$ & $69$ & $207$ & $457$ & $2,285$ \\
        \bottomrule
    \end{tabular}
\end{table}

\clearpage
\section{All Experimental Results} \label{sec:all_experimental_results}

\begin{table}[ht]
    \caption{Elapsed time [s] per Gibbs iteration (mean $\pm$ std)}
    \centering
    \resizebox{\linewidth}{!}{
    \begin{tabular}{lrrrrrr}
        \toprule
             & \multicolumn{3}{l}{$C_s=2$}                                  & \multicolumn{3}{l}{$C_s=4$} \\
        Dataset & Top-down                    & Bottom-up       & \textit{Speedup} & Top-down           & Bottom-up       & \textit{Speedup} \\
        \midrule
        Abalone            & {\boldmath $0.076\pm0.004$} & $1.574\pm0.043$ & $\times21$ & {\boldmath $0.367\pm0.019$} & $17.38\pm0.699$ & $\times47$ \\
        Ailerons           & {\boldmath $2.938\pm0.121$} & $96.28\pm2.631$ & $\times33$ & {\boldmath $181.2\pm28.06$} & $3282\pm105.9$  & $\times18$ \\
        Airfoil Self-Noise & {\boldmath $0.010\pm0.001$} & $0.303\pm0.024$ & $\times30$ & {\boldmath $0.021\pm0.001$} & $2.047\pm0.143$ & $\times97$ \\
        Breast             & {\boldmath $0.005\pm0.000$} & $0.170\pm0.016$ & $\times34$ & {\boldmath $0.014\pm0.001$} & $1.938\pm0.079$ & $\times138$ \\
        Computer Hardware  & {\boldmath $0.003\pm0.000$} & $0.098\pm0.011$ & $\times33$ & {\boldmath $0.007\pm0.000$} & $0.943\pm0.078$ & $\times135$ \\
        cpu\_act            & {\boldmath $0.525\pm0.016$} & $16.80\pm0.771$ & $\times32$ & {\boldmath $11.72\pm2.503$} & $588.0\pm15.46$ & $\times50$ \\
        cpu\_small          & {\boldmath $0.204\pm0.005$} & $5.592\pm0.206$ & $\times27$ & {\boldmath $2.059\pm0.119$} & $99.17\pm3.382$ & $\times48$ \\
        Crx                & {\boldmath $0.023\pm0.001$} & $0.722\pm0.013$ & $\times31$ & {\boldmath $0.232\pm0.006$} & $14.21\pm0.458$ & $\times61$ \\
        Dermatology        & {\boldmath $0.034\pm0.002$} & $1.887\pm0.042$ & $\times56$ & {\boldmath $1.932\pm0.162$} & $92.19\pm3.150$ & $\times48$ \\
        elevators          & {\boldmath $0.759\pm0.025$} & $24.87\pm0.635$ & $\times33$ & {\boldmath $17.62\pm3.630$} & $804.7\pm35.45$ & $\times46$ \\
        Forest Fires       & {\boldmath $0.015\pm0.001$} & $0.405\pm0.021$ & $\times27$ & {\boldmath $0.101\pm0.004$} & $5.809\pm0.191$ & $\times58$ \\
        German             & {\boldmath $0.050\pm0.003$} & $2.095\pm0.064$ & $\times42$ & {\boldmath $1.298\pm0.099$} & $76.78\pm3.338$ & $\times59$ \\
        Housing            & {\boldmath $0.009\pm0.000$} & $0.445\pm0.038$ & $\times49$ & {\boldmath $0.114\pm0.005$} & $6.616\pm0.250$ & $\times58$ \\
        Hybrid Price       & {\boldmath $0.001\pm0.000$} & $0.015\pm0.000$ & $\times15$ & {\boldmath $0.001\pm0.000$} & $0.061\pm0.007$ & $\times61$    \\
        kin8nm             & {\boldmath $0.099\pm0.005$} & $3.216\pm0.326$ & $\times32$ & {\boldmath $0.696\pm0.034$} & $33.76\pm1.653$ & $\times49$ \\
        LPGA2008          & {\boldmath $0.001\pm0.000$} & $0.046\pm0.002$ & $\times46$ & {\boldmath $0.002\pm0.000$} & $0.318\pm0.034$ & $\times159$   \\
        LPGA2009          & {\boldmath $0.002\pm0.000$} & $0.118\pm0.014$ & $\times59$ & {\boldmath $0.011\pm0.001$} & $1.347\pm0.061$ & $\times123$ \\
        Parkinsons Telemonitoring (motor) & {\boldmath $0.207\pm0.012$} & $6.516\pm0.189$ & $\times31$ & {\boldmath $3.273\pm0.300$} & $157.8\pm5.869$ & $\times48$ \\
        Parkinsons Telemonitoring (total) & {\boldmath $0.211\pm0.011$} & $6.631\pm0.226$ & $\times30$ & {\boldmath $3.207\pm0.332$} & $156.4\pm5.911$ & $\times49$ \\
        Vote for Clinton   & {\boldmath $0.028\pm0.002$} & $1.413\pm0.153$ & $\times50$ & {\boldmath $0.315\pm0.011$} & $15.43\pm0.510$ & $\times49$ \\
        Wine Quality Red   & {\boldmath $0.024\pm0.002$} & $1.100\pm0.079$ & $\times46$ & {\boldmath $0.294\pm0.012$} & $15.51\pm0.871$ & $\times53$ \\
        Wine Quality White & {\boldmath $0.088\pm0.006$} & $3.215\pm0.249$ & $\times37$ & {\boldmath $0.979\pm0.044$} & $49.60\pm2.455$ & $\times51$ \\
        Wine               & {\boldmath $0.004\pm0.000$} & $0.196\pm0.020$ & $\times49$ & {\boldmath $0.028\pm0.004$} & $2.576\pm0.088$ & $\times92$ \\
        Yacht Hydrodynamics & {\boldmath $0.003\pm0.000$} & $0.088\pm0.012$ & $\times29$ & {\boldmath $0.005\pm0.000$} & $0.603\pm0.064$ & $\times121$ \\
        \bottomrule
    \end{tabular}
    }
\end{table}

\begin{table}[ht]
    \caption{Effective sample size per 50 samples (mean $\pm$ std)}
    \centering
    \begin{tabular}{lrrrr}
        \toprule
                                        & \multicolumn{2}{l}{$C_s=2$}                 & \multicolumn{2}{l}{$C_s=4$} \\
        \multicolumn{1}{l}{Dataset}     & Top-down & Bottom-up                        & Top-down & Bottom-up \\
        \midrule
        Abalone                         & $19.75\pm0.45$ & {\boldmath $20.39\pm0.40$} & $17.36\pm0.28$             & {\boldmath $19.77\pm0.17$} \\
        Ailerons                        & $14.11\pm0.26$ & {\boldmath $16.74\pm0.18$} & {\boldmath $17.92\pm0.07$} & - \\
        Airfoil Self-Noise              & $20.22\pm0.28$ & {\boldmath $20.47\pm0.55$} & $17.89\pm0.52$             & {\boldmath $20.49\pm0.15$} \\
        Breast                          & $20.37\pm0.13$ & {\boldmath $20.56\pm0.15$} & {\boldmath $20.51\pm0.04$} & $20.40\pm0.04$ \\
        Computer Hardware               & $20.32\pm0.43$ & {\boldmath $20.69\pm0.24$} & $19.52\pm0.24$             & {\boldmath $20.57\pm0.03$} \\
        cpu\_act                         & $14.37\pm0.63$ & {\boldmath $20.23\pm0.22$} & {\boldmath $15.50\pm0.70$} & - \\
        cpu\_small                       & $12.94\pm0.99$ & {\boldmath $19.71\pm0.33$} & $17.79\pm0.16$ & {\boldmath $20.18\pm0.13$} \\
        Crx                             & $18.67\pm0.64$ & {\boldmath $20.63\pm0.13$} & $18.84\pm0.07$ & {\boldmath $19.93\pm0.03$} \\
        Dermatology                     & $15.56\pm0.25$ & {\boldmath $20.57\pm0.06$} & $19.07\pm0.06$ & {\boldmath $20.10\pm0.03$} \\
        elevators                       & $15.05\pm0.22$ & {\boldmath $18.01\pm0.34$} & {\boldmath $16.99\pm0.06$} & - \\
        Forest Fires                    & $18.86\pm0.45$ & {\boldmath $20.61\pm0.18$} & $18.60\pm0.11$ & {\boldmath $20.53\pm0.03$} \\
        German                          & $17.34\pm0.34$ & {\boldmath $20.60\pm0.11$} & $18.94\pm0.07$ & {\boldmath $19.35\pm0.03$} \\
        Housing                         & $17.31\pm1.22$ & {\boldmath $20.56\pm0.14$} & $17.43\pm0.24$ & {\boldmath $20.50\pm0.02$} \\
        Hybrid Price                    & $20.38\pm0.66$ & {\boldmath $20.54\pm0.97$} & $20.51\pm0.39$ & {\boldmath $20.57\pm0.36$} \\
        kin8nm                          & $18.15\pm2.14$ & {\boldmath $20.75\pm0.30$} & $16.83\pm0.13$ & {\boldmath $19.45\pm0.18$} \\
        LPGA2008                       & {\boldmath $20.67\pm0.60$} & $20.46\pm0.24$ & $20.47\pm0.17$ & {\boldmath $20.50\pm0.11$} \\
        LPGA2009                       & {\boldmath $20.52\pm0.39$} & $20.48\pm0.31$ & $20.37\pm0.15$ & {\boldmath $20.54\pm0.05$} \\
        Parkinsons Telemonitoring (motor) & $16.43\pm2.41$ & {\boldmath $20.29\pm0.29$} & {\boldmath $17.48\pm0.36$} & $17.07\pm0.06$ \\
        Parkinsons Telemonitoring (total) & $16.54\pm2.52$ & {\boldmath $20.19\pm0.36$} & {\boldmath $18.46\pm0.62$} & $16.96\pm0.10$ \\
        Vote for Clinton                & $16.97\pm0.93$ & {\boldmath $19.95\pm0.58$} & $17.25\pm0.26$ & {\boldmath $19.41\pm0.10$} \\
        Wine Quality Red                & $17.27\pm2.25$ & {\boldmath $20.43\pm0.31$} & $17.04\pm0.16$ & {\boldmath $19.72\pm0.05$} \\
        Wine Quality White              & $15.81\pm1.21$ & {\boldmath $20.54\pm0.29$} & $16.70\pm0.14$ & {\boldmath $19.37\pm0.10$} \\
        Wine                            & $20.51\pm0.35$ & {\boldmath $20.53\pm0.13$} & $18.12\pm0.30$ & {\boldmath $20.55\pm0.03$} \\
        Yacht Hydrodynamics             & $20.03\pm1.16$ & {\boldmath $20.69\pm0.29$} & $20.19\pm0.21$ & {\boldmath $20.57\pm0.11$} \\
        \bottomrule
    \end{tabular}
\end{table}

\begin{table}[ht]
    \caption{Log-likelihood (mean $\pm$ std)}
    \centering
    \resizebox{\linewidth}{!}{
    \begin{tabular}{lrrrr}
        \toprule
                                        & \multicolumn{2}{l}{$C_s=2$}                     & \multicolumn{2}{l}{$C_s=4$} \\
        Dataset                         & Top-down & Bottom-up                            & Top-down & Bottom-up \\
        \midrule
        Abalone                         & {\boldmath $4.47\pm0.31$} & $1.10\pm0.15$       & {\boldmath $6.97\pm0.30$} & $2.94\pm0.12$ \\
        Ailerons                        & {\boldmath $110.24\pm0.87$} & $108.22\pm0.36$   & {\boldmath $106.95\pm0.83$} & $106.26\pm0.99$ \\
        Airfoil Self-Noise              & {\boldmath $-12.57\pm1.25$} & $-13.06\pm1.39$   & {\boldmath $-6.57\pm0.20$} & $-7.91\pm0.27$ \\
        Breast                          & {\boldmath $-6.56\pm0.18$} & $-7.29\pm0.32$     & {\boldmath $-6.45\pm0.11$} & $-6.90\pm0.16$ \\
        Computer Hardware               & {\boldmath $-49.19\pm4.61$} & $-51.94\pm5.76$   & {\boldmath $-32.35\pm0.71$} & $-34.28\pm0.64$ \\
        cpu\_act                         & {\boldmath $-101.13\pm1.49$} & $-103.49\pm1.03$ & $-99.91\pm0.13$ & {\boldmath $-97.79\pm0.20$} \\
        cpu\_small                       & {\boldmath $-85.03\pm0.76$} & $-87.06\pm2.03$   & $-82.52\pm0.62$ & {\boldmath $-82.09\pm0.32$} \\
        Crx                             & {\boldmath $-49.87\pm11.20$} & $-52.13\pm13.01$ & $-30.33\pm2.23$ & {\boldmath $-30.08\pm2.18$} \\
        Dermatology                     & $21.87\pm2.99$ & {\boldmath $22.26\pm1.57$}     & $25.92\pm0.55$ & {\boldmath $46.70\pm0.26$} \\
        elevators                       & {\boldmath $29.07\pm0.15$} & $29.02\pm0.14$     & {\boldmath $29.34\pm0.16$} & $29.14\pm0.19$ \\
        Forest Fires                    & {\boldmath $-30.81\pm0.49$} & $-32.24\pm0.60$   & {\boldmath $-27.46\pm0.25$} & $-28.45\pm0.15$ \\
        German                          & {\boldmath $-17.52\pm0.45$} & $-19.00\pm0.32$   & $-15.33\pm0.60$  & {\boldmath $-14.08\pm0.23$} \\
        Housing                         & {\boldmath $-24.07\pm1.33$} & $-27.02\pm0.95$   & {\boldmath $-18.40\pm0.94$} & $-20.80\pm0.21$ \\
        Hybrid Price                    & {\boldmath $-22.58\pm0.42$} & $-22.74\pm0.29$   & {\boldmath $-21.42\pm0.36$} & $-22.61\pm0.19$ \\
        kin8nm                          & {\boldmath $-10.23\pm0.10$} & $-10.26\pm0.07$   & {\boldmath $-10.10\pm0.04$} & $-10.13\pm0.02$ \\
        LPGA2008                       & {\boldmath $-15.68\pm0.42$} & $-15.93\pm0.43$   & {\boldmath $-14.68\pm0.20$} & $-14.84\pm0.09$ \\
        LPGA2009                       & {\boldmath $-28.29\pm0.40$} & $-29.63\pm0.21$   & {\boldmath $-27.39\pm0.44$} & $-29.72\pm0.20$ \\
        Parkinsons Telemonitoring (motor) & {\boldmath $43.85\pm0.65$} & $41.94\pm0.25$     & {\boldmath $43.50\pm0.56$} & $42.08\pm0.06$ \\
        Parkinsons Telemonitoring (total) & {\boldmath $43.37\pm0.57$} & $41.68\pm0.13$     & {\boldmath $42.82\pm0.83$} & $41.53\pm0.07$ \\
        Vote for Clinton                & {\boldmath $-49.61\pm0.14$} & $-49.84\pm0.03$   & {\boldmath $-49.77\pm0.09$} & $-49.89\pm0.05$ \\
        Wine Quality Red                & {\boldmath $-3.62\pm0.30$} & $-4.35\pm0.37$     & {\boldmath $-2.73\pm0.28$} & $-2.95\pm0.11$ \\
        Wine Quality White              & {\boldmath $-2.96\pm0.57$} & $-3.59\pm0.56$     & {\boldmath $-2.57\pm0.14$} & $-2.61\pm0.10$ \\
        Wine                            & {\boldmath $-19.90\pm0.63$} & $-22.44\pm0.46$   & {\boldmath $-18.41\pm0.44$} & $-20.58\pm0.14$ \\
        Yacht Hydrodynamics             & {\boldmath $2.22\pm1.83$} & $1.06\pm1.85$       & {\boldmath $8.25\pm0.64$} & $4.98\pm0.15$ \\
        \bottomrule
    \end{tabular}
    }
\end{table}

\begin{figure}[h]
    \includegraphics[keepaspectratio, width=0.24\linewidth]{240613-llh_legend.png}
    \includegraphics[keepaspectratio, width=0.24\linewidth]{240613-llh_abalone.png}
    \includegraphics[keepaspectratio, width=0.24\linewidth]{240613-llh_ailerons.png}
    \includegraphics[keepaspectratio, width=0.24\linewidth]{240613-llh_airfoil-self-noise.png}
    \\
    \includegraphics[keepaspectratio, width=0.24\linewidth]{240613-llh_breast-cancer.png}
    \includegraphics[keepaspectratio, width=0.24\linewidth]{240613-llh_computer-hardware.png}
    \includegraphics[keepaspectratio, width=0.24\linewidth]{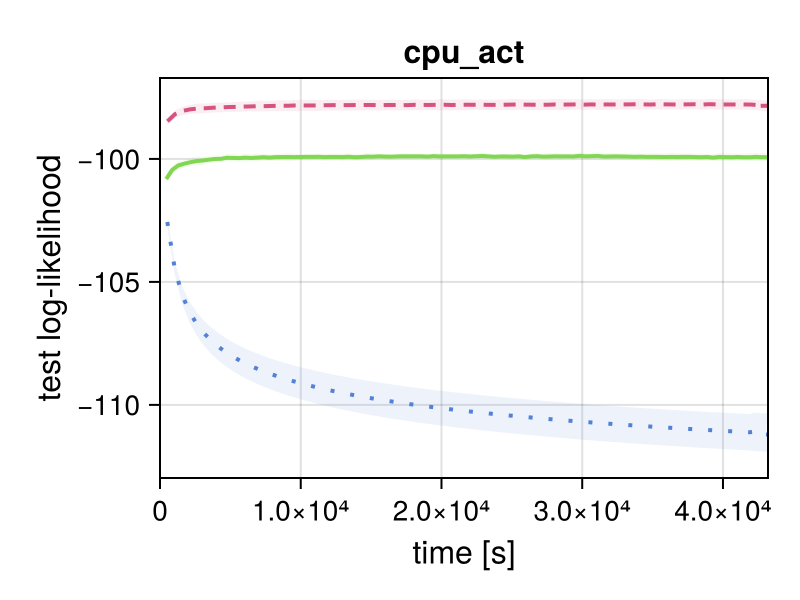}
    \includegraphics[keepaspectratio, width=0.24\linewidth]{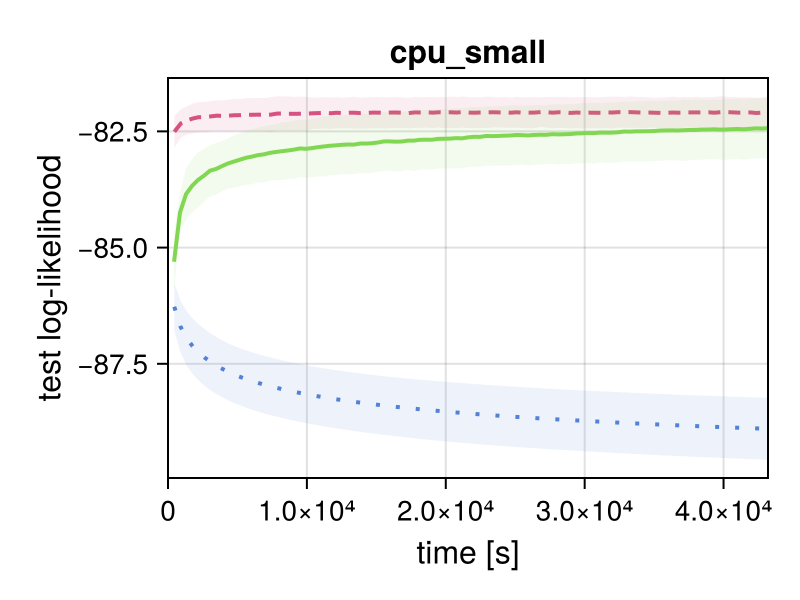}
    \\
    \includegraphics[keepaspectratio, width=0.24\linewidth]{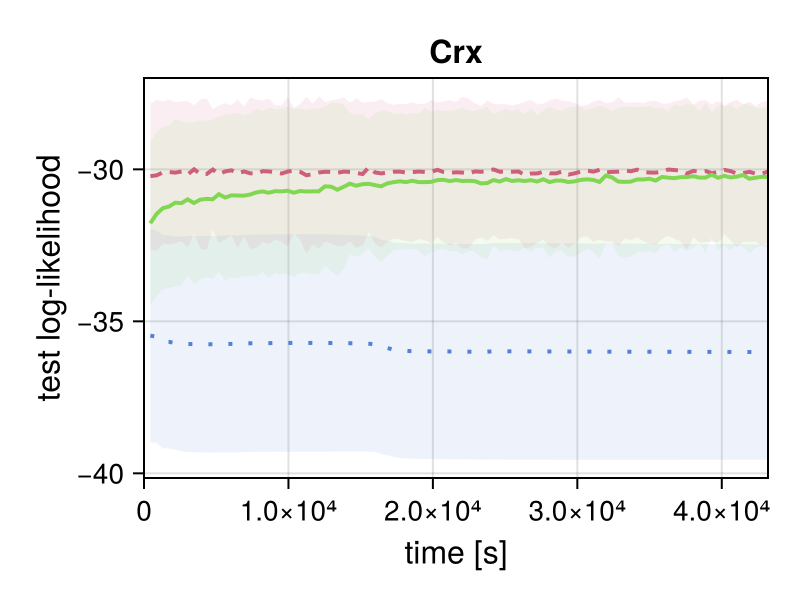}
    \includegraphics[keepaspectratio, width=0.24\linewidth]{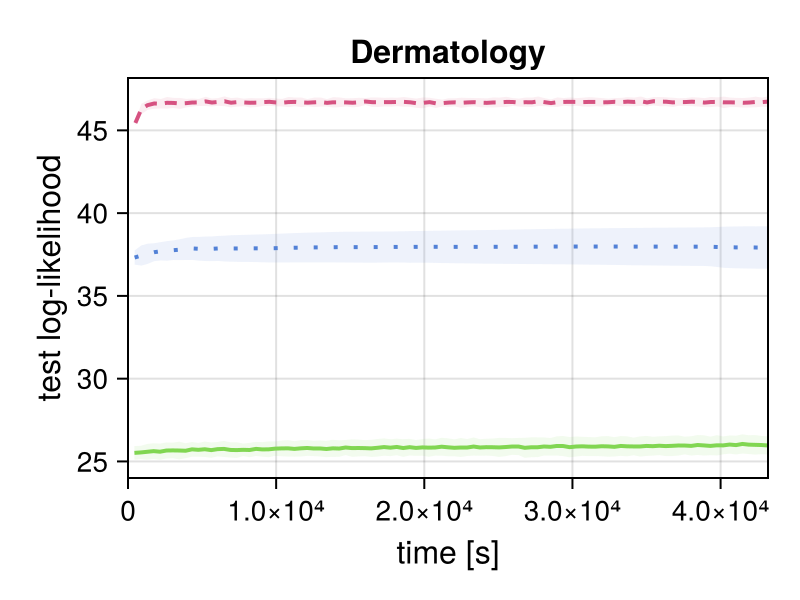}
    \includegraphics[keepaspectratio, width=0.24\linewidth]{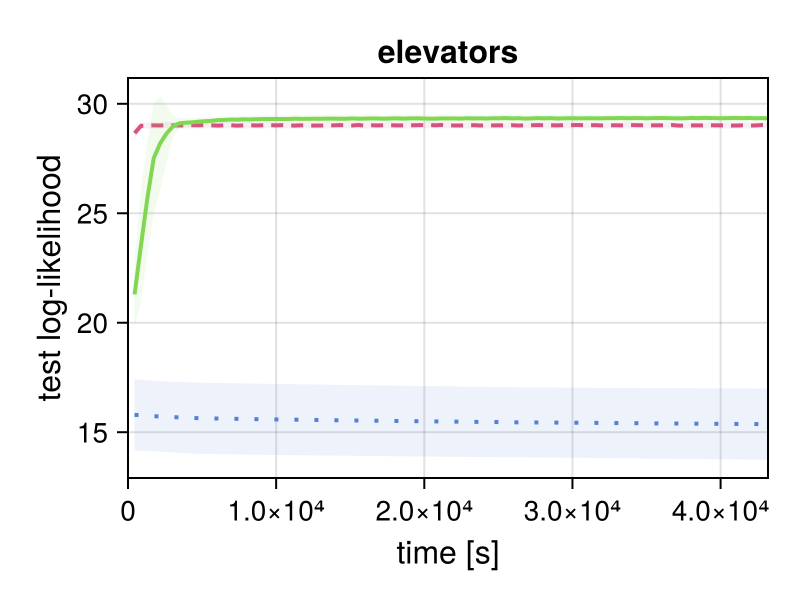}
    \includegraphics[keepaspectratio, width=0.24\linewidth]{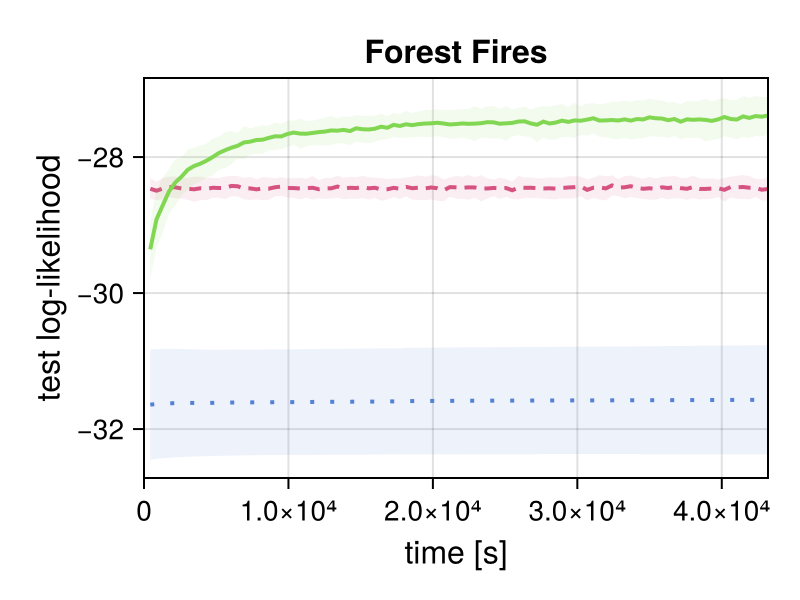}
    \\
    \includegraphics[keepaspectratio, width=0.24\linewidth]{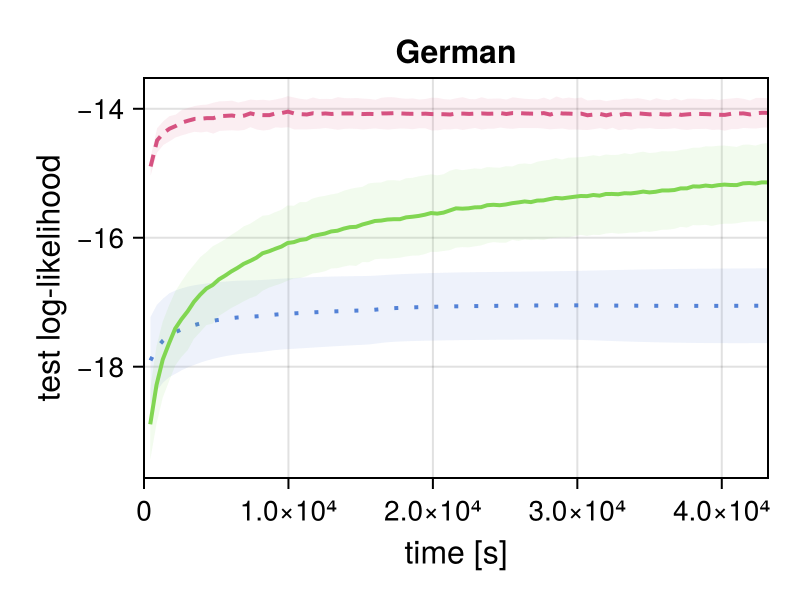}
    \includegraphics[keepaspectratio, width=0.24\linewidth]{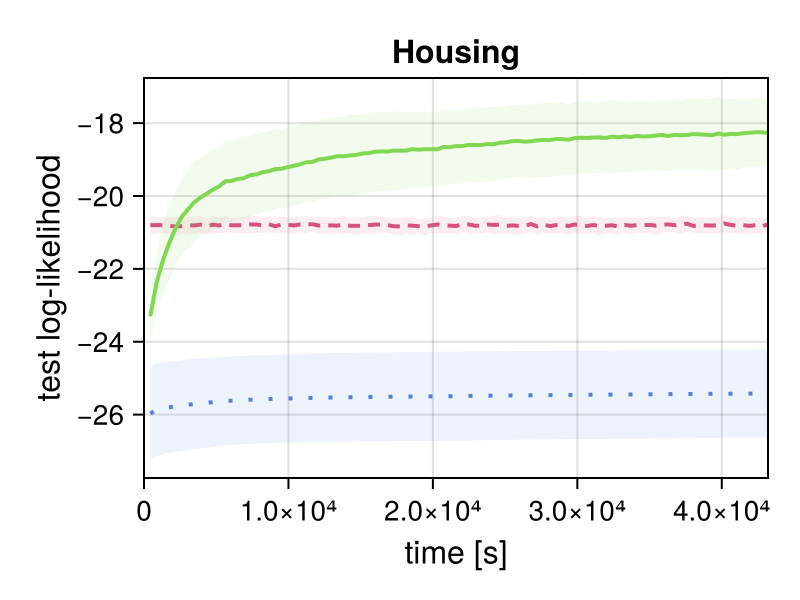}
    \includegraphics[keepaspectratio, width=0.24\linewidth]{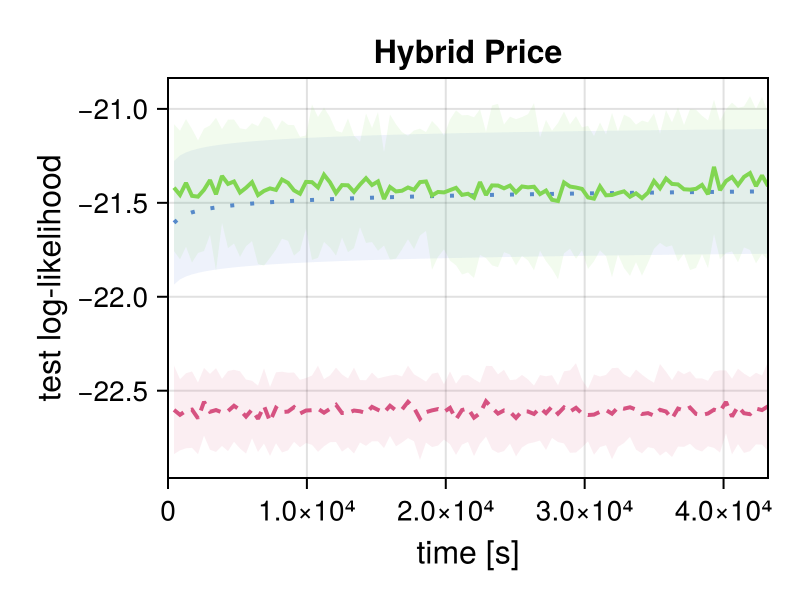}
    \includegraphics[keepaspectratio, width=0.24\linewidth]{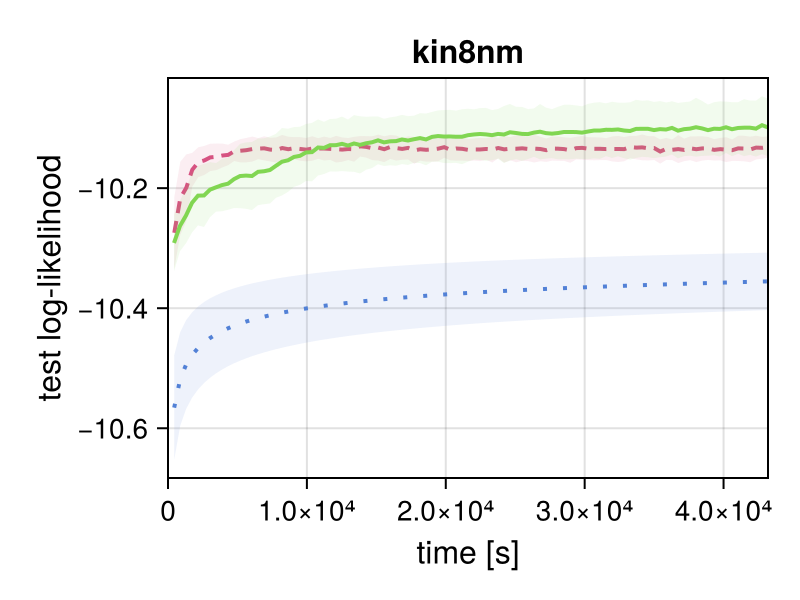}
    \\
    \includegraphics[keepaspectratio, width=0.24\linewidth]{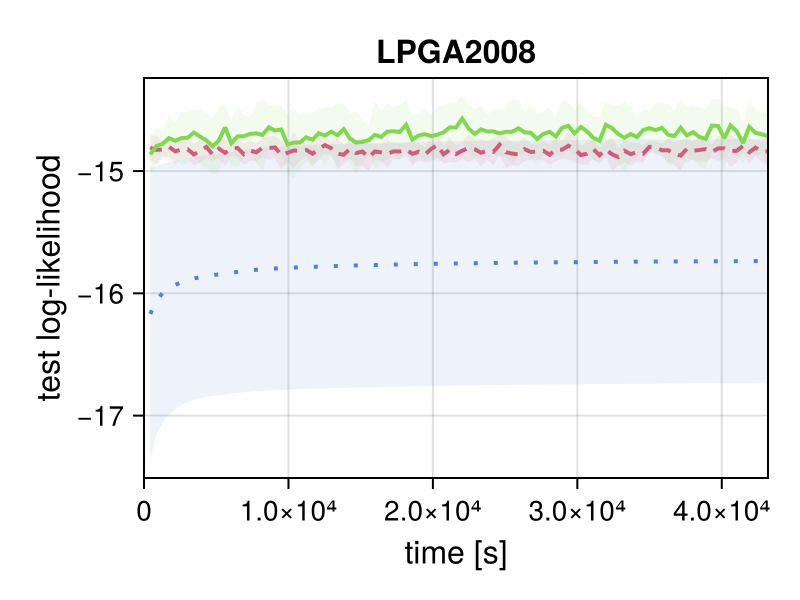}
    \includegraphics[keepaspectratio, width=0.24\linewidth]{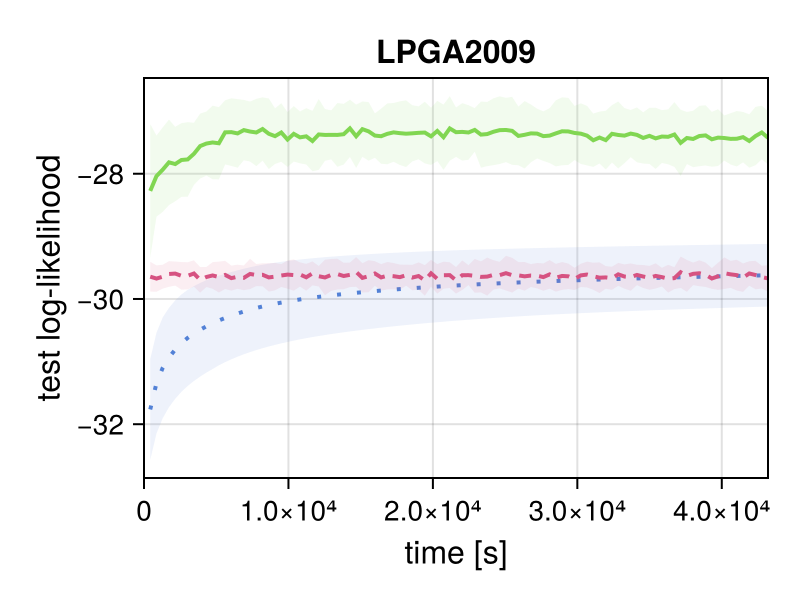}
    \includegraphics[keepaspectratio, width=0.24\linewidth]{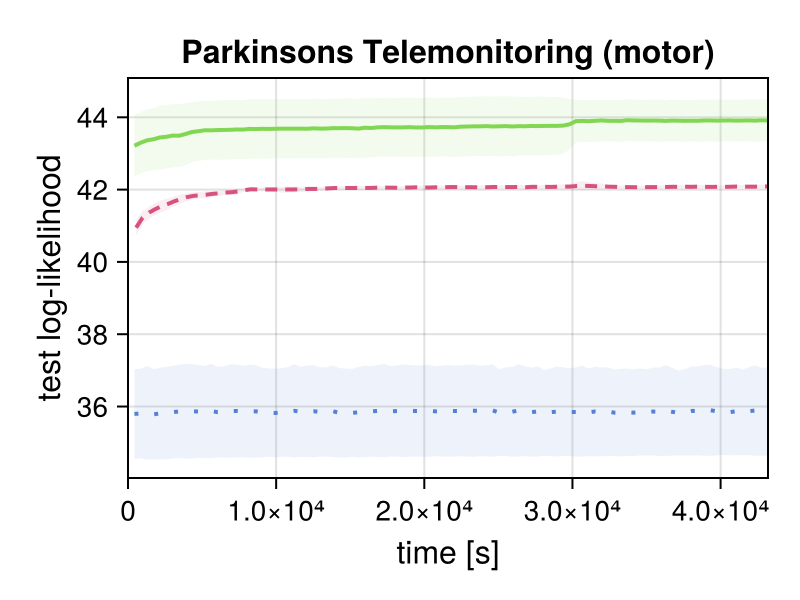}
    \includegraphics[keepaspectratio, width=0.24\linewidth]{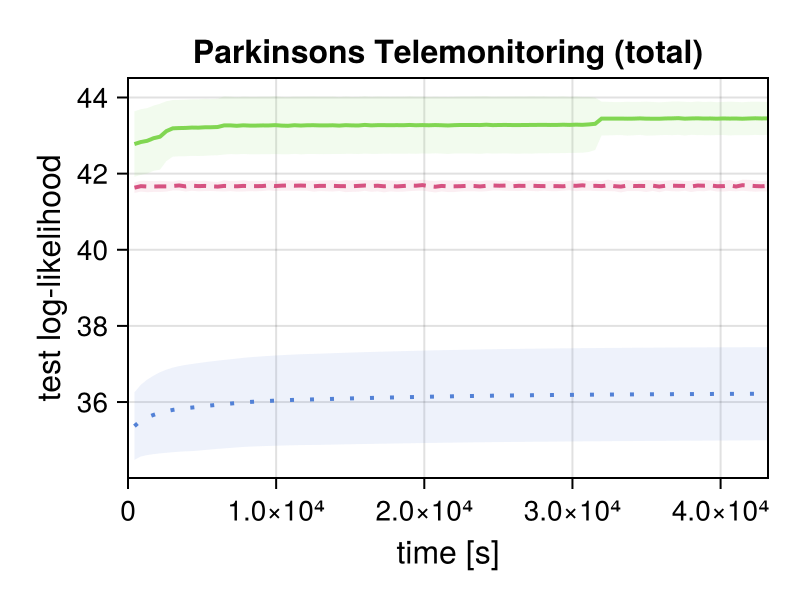}
    \\
    \includegraphics[keepaspectratio, width=0.24\linewidth]{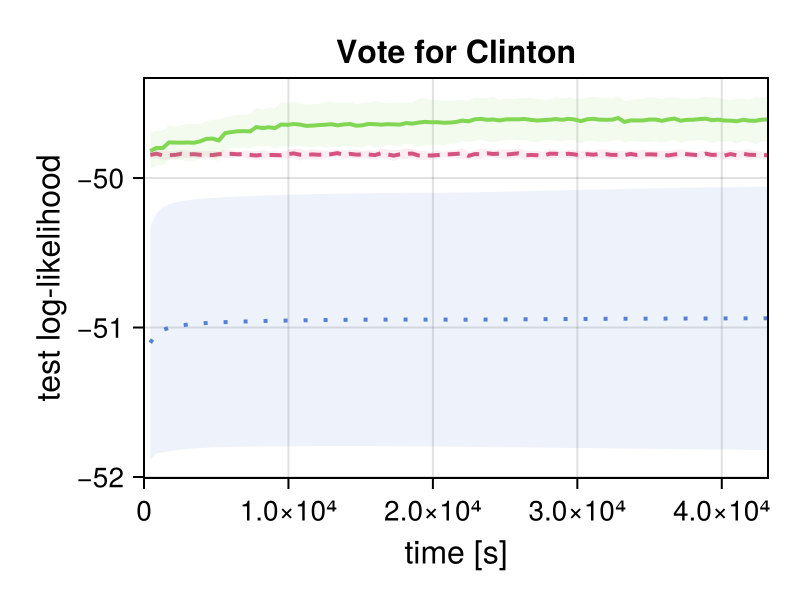}
    \includegraphics[keepaspectratio, width=0.24\linewidth]{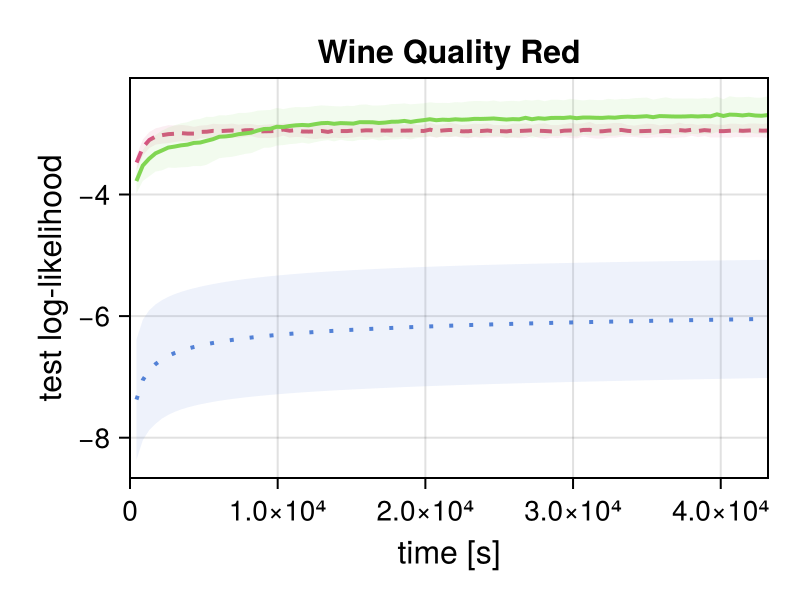}
    \includegraphics[keepaspectratio, width=0.24\linewidth]{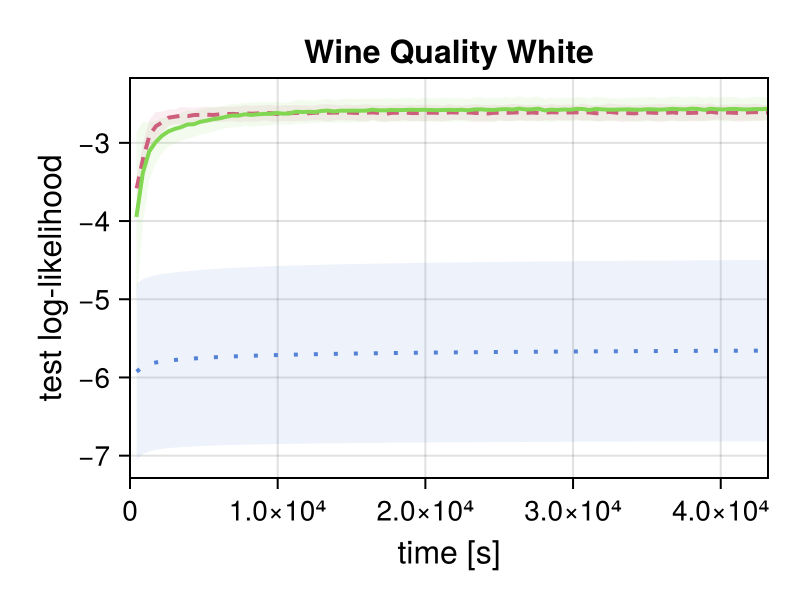}
    \includegraphics[keepaspectratio, width=0.24\linewidth]{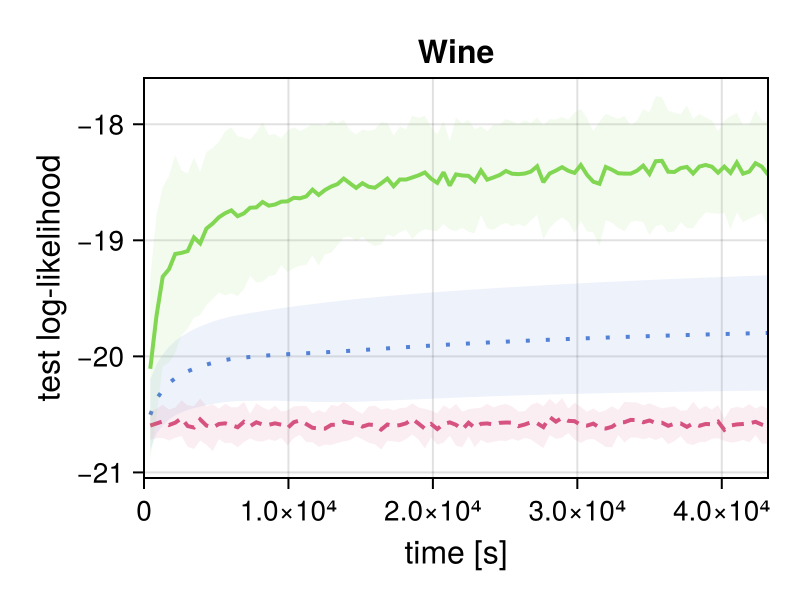}
    \\
    \includegraphics[keepaspectratio, width=0.24\linewidth]{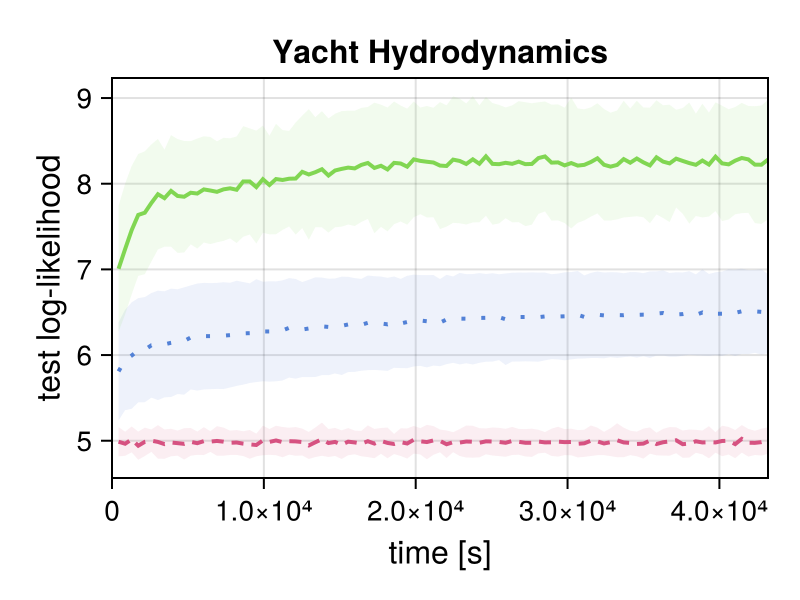}
    \caption{Temporal evolution in predictive performance for each dataset. The average test-set log-likelihood over $10$ trials is shown with the lines, and the standard deviation is indicated by the shaded regions.}
    \label{fig:SPN_exp_llh}
\end{figure}

\end{document}